\documentclass[10pt,twocolumn,letterpaper]{article}

\usepackage{cvpr}              
\usepackage{amsmath}
\usepackage{cite}
\usepackage{float}
\usepackage{afterpage}
\usepackage{supertabular}
\usepackage{lipsum}
\usepackage{makecell}       
\usepackage{algorithm,algorithmicx,algpseudocode}
\usepackage{algorithm,algpseudocode}
\usepackage{graphicx}
\usepackage{booktabs}
\usepackage{multirow}
\usepackage{mathtools}
\usepackage[labelfont=bf]{caption}

\usepackage[table,dvipsnames]{xcolor}
\usepackage{booktabs}
\usepackage{multirow}
\usepackage{arydshln} 
\usepackage{graphicx}
\usepackage{array}
\usepackage{rotating}
\usepackage[export]{adjustbox}
\usepackage{colortbl}

\usepackage{nicematrix}

\definecolor{cvprblue}{rgb}{0.21,0.49,0.74}
\usepackage[pagebackref,breaklinks,colorlinks,citecolor=cvprblue]{hyperref}

\def\paperID{419} 
\def\confName{3DV\xspace}
\def\confYear{2026\xspace}

\title{Proxy-Free Gaussian Splats Deformation with Splat-Based Surface Estimation}

\author{
Jaeyeong Kim\qquad
Seungwoo Yoo\qquad
Minhyuk Sung\\
KAIST\\
{\tt\small jy9394@kaist.ac.kr \quad dreamy1534@kaist.ac.kr \quad mhsung@kaist.ac.kr}\\
}

\begin{document}
\twocolumn[{%
  \maketitle
  \begin{center}
    \includegraphics[width=\linewidth]{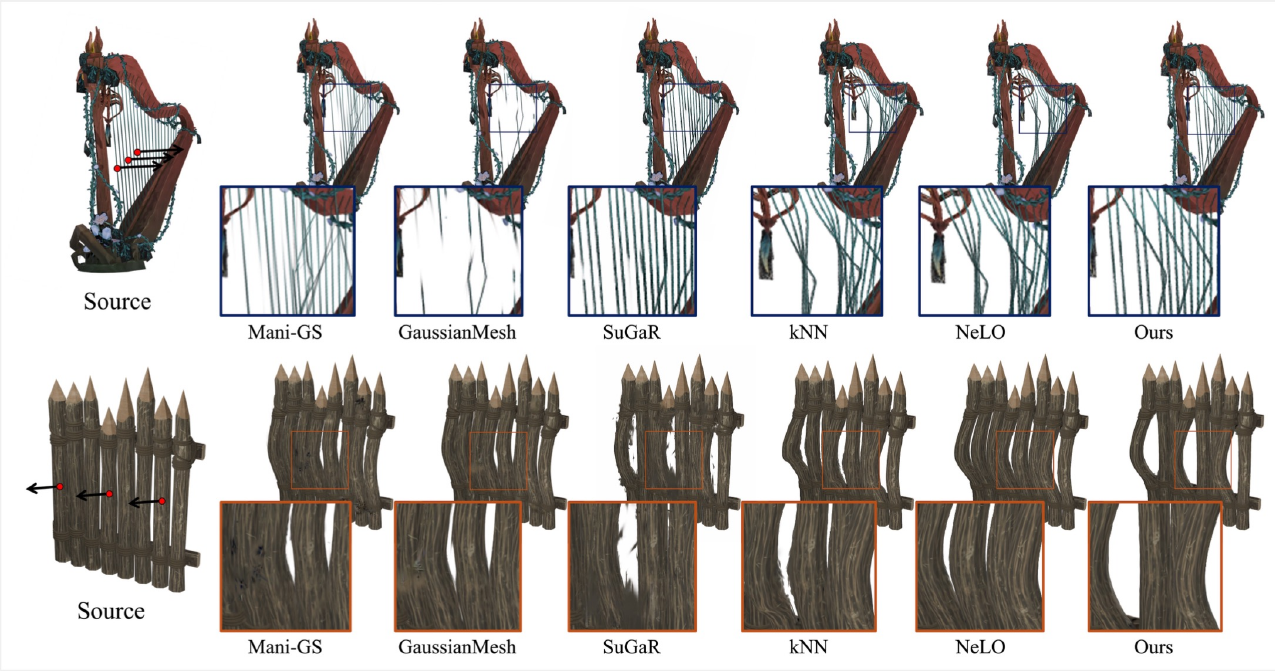}
    \captionof{figure}{\textbf{SpLap} is proxy-free Laplacian-based deformation framework for Gaussian splats. 
    Our method enables high-quality, large-scale deformation, all without relying on external proxy.
    We show results for both As-Rigid-As-Possible~\cite{ARAP} (top row) and Bounded Biharmonic Weights~\cite{BBW} (bottom row) deformations. The red dots indicate the interaction handles and arrows show the direction of the applied edit.}
    \label{fig:teaser}
  \end{center}
}]
\maketitle
\begin{abstract}
We introduce SpLap, a proxy-free deformation method for Gaussian splats (GS) based on a Laplacian operator computed from our novel surface-aware splat graph. 
Existing approaches to GS deformation typically rely on deformation proxies such as cages or meshes, but they suffer from dependency on proxy quality and additional computational overhead.
An alternative is to directly apply Laplacian-based deformation techniques by treating splats as point clouds. 
However, this often fail to properly capture surface information due to lack of explicit structure.
To address this, we propose a novel method that constructs a surface-aware splat graph, enabling the Laplacian operator derived from it to support more plausible deformations that preserve details and topology. 
Our key idea is to leverage the spatial arrangement encoded in splats, defining neighboring splats not merely by the distance between their centers, but by their intersections.
Furthermore, we introduce a Gaussian kernel adaptation technique that preserves surface structure under deformation, thereby improving rendering quality after deformation. 
In our experiments, we demonstrate the superior performance of our method compared to both proxy-based and proxy-free baselines, evaluated on 50 challenging objects from the ShapeNet, Objaverse, and Sketchfab datasets, as well as the NeRF-Synthetic dataset. Code is available at \url{https://github.com/kjae0/SpLap}.
\end{abstract}
    
\section{Introduction}
\label{sec:intro}
Deforming 3D shapes has long been a central problem in geometry processing. 
For decades, extensive research has provided well-established deformation techniques for 3D representations, most notably polygonal meshes~\cite{ARAP, BBW, APAP}. 
The goal of these techniques is to allow user-driven manipulation while preserving the intrinsic geometry of the shape. 
As a representative approach, Laplacian-based methods achieve this by minimizing a deformation energy over the geometry to penalize undesirable distortions.

Recently, 3D Gaussian Splatting (3DGS)~\cite{kerbl3Dgaussians} has shown photorealistic novel view synthesis (NVS) quality at real-time frame rates. 
Although the explicit point-based nature of 3DGS primitives suggests a potential for direct manipulation, it often leads to suboptimal results due to the lack of connectivity and structural prior. 
While previous efforts have proposed deformation methods that learn a motion prior from the dynamic scene~\cite{huang2023sc, waczyńska2024dmiso, yao2025riggs, yang2023deformable3dgs}, these are not applicable to a general task of editing static objects. 
Consequently, the prevalent workaround is to introduce a deformable geometric proxy, such as a mesh~\cite{waczynska2024games, MeshGaussian2024, gao2024mani} or cage~\cite{jiang2024vr-gs, zielonka25dega, huang2024gsdeformerdirectrealtimeextensible, cagegs}. These methods parametrize Gaussian kernels with respect to a proxy such that manipulations of the proxy deform the associated Gaussian kernels.
    
However, this reliance on a proxy introduces two critical limitations: the final deformation quality is fundamentally limited by the quality of the proxy. 
In the case of meshes, which are the most widely used as a proxy, geometric defects—such as topological errors or missing parts—propagate directly into the deformation, causing severe visual artifacts and narrowing the applicability of existing deformation techniques. 
Moreover, this strategy incurs an additional computational cost, which acts as a practical barrier. 
This approach necessitates a lengthy pipeline of mesh generation, where high-quality surface reconstruction such as NeuS~\cite{wang2021neus} or Neuralangelo~\cite{li2023neuralangelo} takes at least several GPU hours, often followed by laborious post-processing to ensure the mesh is suitable for deformation.

In this paper, to remedy these limitations derived from the dependency on proxy, we propose SpLap, a proxy-free Laplacian-based deformation framework for Gaussian splats. Rather than relying on an external structural prior, our approach leverages an inherent surface geometry encoded within the spatial arrangement of the surface-aligned Gaussian splats~\cite{Huang2DGS2024, zhang2024spiking, Dai2024GaussianSurfels, guedon2023sugar}. Our key observation is that a surface-aligned GS can be interpreted as a discrete, patch-based surface representation, since each primitive is optimized to act as a local chart of the manifold. 
Motivated by this observation, we leverage this inherent surface alignment of primitives to extract geometric information that is crucial to utilize traditional deformation techniques~\cite{ARAP, BBW}.

Our framework begins with establishing the connectivity from the Gaussian kernels. 
However, since simple spatial proximity often fails to capture a valid adjacency on the surface manifold, we incorporate intrinsic surface information by leveraging surface-aligned GS. 
Specifically, we construct a splat intersection graph whose connectivity is defined by the overlap of Gaussian kernels.
This intersection-based connectivity is the key component that allows us to estimate the underlying surface without an external geometric proxy.
Subsequently, we directly build Laplacian operator and apply Laplacian-based geometric deformation techniques to Gaussian kernels, treating their means as a point cloud.
By building a Laplacian operator upon our surface-aware splat intersection graph, we overcome the limitations of existing point cloud Laplacian methods and enable topologically sound deformations even on complex geometries such as adjacent distinct surfaces or thin regions.
Finally, we introduce an adaptation method that is crucial for preserving visual fidelity. 
As the displaced Gaussian kernels no longer reflect the deformed local surface geometry, we adapt each primitive to maintain the original coverage on the surface manifold. 
Consequently, our framework enables direct geometric deformation for Gaussian splats while preserving photorealistic quality even under a large-scale deformation.

In experiments, we demonstrate the robustness and efficiency of our approach through experiments on a diverse set of objects with intricate geometries. 
Applying two representative Laplacian-based deformation techniques, As-Rigid-As-Possible (ARAP)~\cite{ARAP} and Bounded Biharmonic Weights (BBW)~\cite{BBW}, we show that our method achieves a deformation fidelity nearly indistinguishable from the ground truth result, whereas existing methods fail to produce a plausible quality.


\section{Related Works}
\label{sec:formatting}

\subsection{Gaussian Splats Deformation}
Diverse strategies for Gaussian splats deformation have been explored including editable dynamic scenes~\cite{huang2023sc, waczyńska2024dmiso, yang2023deformable3dgs, yao2025riggs}, simulation-based methods~\cite{xie2023physgaussian, jiang2024vr-gs}.
However, these methods are not applicable to a general task of direct, user-guided editing for static objects.
Consequently, methods that leverage geometric templates such as a mesh or cage as a deformable proxy, have been proposed~\cite{waczynska2024games, MeshGaussian2024, gao2024mani, huang2024gsdeformerdirectrealtimeextensible, guedon2023sugar, guedon2024frosting, zielonka25dega}. Although GSDeformer~\cite{huang2024gsdeformerdirectrealtimeextensible} and Cage-GS~\cite{cagegs} propose a cage-based deformation approach, these lack fine-grained control. For this reason, mesh-based approaches have become an attractive alternative, providing detailed control over the deformation. Specifically, SuGaR~\cite{guedon2023sugar} and GaussianFrosting~\cite{guedon2024frosting} extract a mesh directly from Gaussian primitives via Poisson surface reconstruction~\cite{kazhdan2006poisson} and then parametrize the primitives to the mesh. Otherwise, Mani-GS~\cite{gao2024mani} and GaussianMesh~\cite{MeshGaussian2024} generate templates through a separate reconstruction stage using NeuS~\cite{wang2021neus} and NeuS2~\cite{neus2} and bind the Gaussian kernels to the mesh faces. However, the effectiveness of these mesh-based methods is fundamentally limited by the quality of the template mesh. Although Mani-GS~\cite{gao2024mani} introduces learnable offsets to compensate for a defective template, this cannot resolve the inaccurate deformations that arise from a flawed geometric proxy. 
In contrast to these previous works, our approach does not rely on any explicit geometric proxy while enabling direct and point-level deformation.
\subsection{Gaussian Splatting with Surface Geometry}
Recently, there have been efforts to integrate a surface geometry to Gaussian Splatting~\cite{kerbl3Dgaussians}, while keeping photorealistic rendering~\cite{yu2024gsdf, 3dgsr, Yu2024GOF, chen2025neusgneuralimplicitsurface}. For instance, several works directly regularize the Gaussian primitives to be flattened and aligned with the underlying surface~\cite{guedon2023sugar, guedon2024frosting}. Inspired by earlier surfel-based representations~\cite{surfel:surface, surfel_splatting, EWA_surfel}, 2DGS~\cite{Huang2DGS2024} and G-Surfel~\cite{Dai2024GaussianSurfels} even replace 3D primitives with 2D Gaussian kernels with additional geometric regularizations, such as depth distortion loss~\cite{barron2022mipnerf360} and normal consistency loss. Built upon 2DGS, SpikingGS~\cite{zhang2024spiking} utilizes spiking neurons with an adaptive opacity threshold, achieving superior surface alignment capability. These works reveal the potential of Gaussian splats can serve as a patch-based discrete surface representation. In this work, we employ this insight for a geometric deformation of Gaussian splats without relying on any external geometric proxy.


\subsection{Laplacian Operator on Point-based Representations}
The discrete Laplacian operator is a fundamental tool for geometry processing~\cite{Botsch2010PolygonMeshProcessing, LBOshapeanalysis}. 
Without explicit connectivity, existing methods typically rely on a $k$-nearest-neighbor (kNN) search for local neighborhood estimation. Based on the kNN graph, various approaches have been proposed including kernel-based methods on triangulation~\cite{belkin09, pointbasedmanifoldharmonics, pcdlap_approx_grad, aniso}, mesh Laplacian schemes~\cite{finiteelempcdsurf, pcdskeletonlap, nonmanifold}, smoothed particle hydrodynamics (SPH)~\cite{SPH} and moving-least-squares (MLS)~\cite{geometry_understanding}. More recently, NeLO~\cite{pang2024nelo} introduced a learning-based approach, using a graph neural network on the kNN graph. Despite these advances, existing methods inherit the surface-agnostic nature of the initial kNN graph and therefore fail to identify the local surface manifold on intricate geometry. 
For Gaussian Splatting, LBO-GS~\cite{zhou2025laplacebeltramioperatorgaussiansplatting} utilizes the Mahalanobis distance to employ additional geometric information of Gaussian kernels. However, this approach is critically sensitive to highly anisotropic primitives and is insufficient for robust surface estimation. In contrast, our work addresses this by introducing an intersection-based surface estimation, thereby enabling robust and plausible Laplacian-based deformation.

\section{Preliminary}
\paragraph{Surface-Aligned Gaussian Splatting.} 
3D Gaussian Splatting~\cite{kerbl3Dgaussians} represents a 3D scene with a set of 3D Gaussian kernels. Each Gaussian kernel $\mathcal{G}$ is parametrized as:
\begin{equation}
\mathcal{G}(\mathbf{x}) = e^{-\frac{1}{2}(\mathbf{x} - \mathbf{p})^\top \Sigma^{-1}(\mathbf{x} - \mathbf{p})}
\end{equation}
where $\mathbf{p}\in \mathbb{R}^3$ and $\Sigma \in \mathbb{R}^{3\times3}$ represent position and covariance matrix. The covariance matrix $\Sigma = \mathbf{RS}\mathbf{S}^{\top}\mathbf{R}^{\top}$ is factorized into a rotation matrix $\mathbf{R} \in \mathbb{R}^{3\times3}$ and diagonal scale matrix $\mathbf{S} \in \mathbb{R}^{3\times3}$. 
For 2D primitive-based methods~\cite{Huang2DGS2024, Dai2024GaussianSurfels, zhang2024spiking}, kernels are simplified into flat 2D Gaussian kernels by fixing the last diagonal element of $\mathbf{S}$ to zero. Typically, these primitives are optimized with photometric losses alongside geometric objectives, such as depth distortion loss~\cite{barron2022mipnerf360} and normal consistency loss. These geometric constraints encourage surface alignment by concentrating the weight distribution of kernels on the underlying surface.

\begin{figure}[t]
  \centering
  \includegraphics[width=\linewidth]{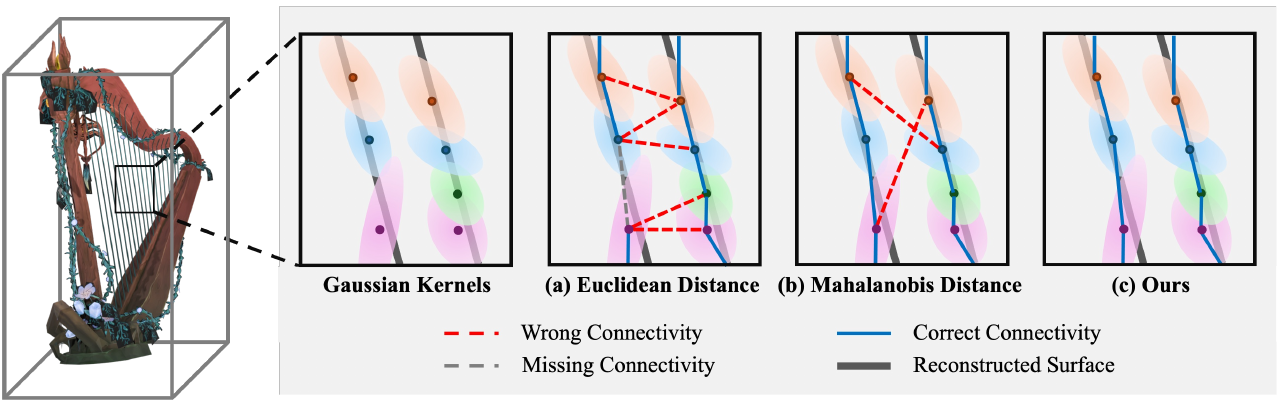}
  \caption{
  \textbf{Illustration of different connectivity search methods.} Existing metrics, (a) and (b), often produce incorrect connections (red) or fragmentation (gray) due to the lack of structural prior. In contrast, by leveraging geometric constraints of surface-aligned GS, our method avoids this suboptimal situations like (c).
  }
  \label{fig:main}
\end{figure}

\section{Method}
\subsection{Method Overview}
The explicit point-based nature of Gaussian Splatting makes the Gaussian kernels amenable to direct manipulation. 
By treating them as a point cloud, the Laplacian operator can be directly defined over the splats using point cloud Laplacian method such as NonManifold~\cite{nonmanifold}.
Although this approach enables Laplacian-based deformation on Gaussian splats, the lack of geometric prior often leads to suboptimal results due to the wrong surface estimation.
Since point cloud Laplacian rely on the connectivity of graph over the nodes, the fidelity of its application is determined by the quality of neighbor estimation.
To handle this, in this work, we present \textbf{SpLap}, a proxy-free deformation framework for Gaussian splats. 
Specifically, our approach leverages surface-aligned GS to extract geometric information directly from the spatial arrangement of the splats themselves. This splat-based surface estimation obviates the need for a geometric proxy, thereby avoiding the limitations of proxy-based methods. 
Specifically, our pipeline is divided into 3 stages:

\begin{enumerate}
    \item \textbf{Surface-Aware Splat Graph Construction (Sec.~\ref{method:SA-graph}):} 
    We first construct a splat graph, where the connectivity is defined by the overlap of splats.
    Based on this graph, we then estimate the surface geometry through a geodesic distance.
    \item \textbf{Laplacian-Based Deformation (Sec.~\ref{method:Deformation_stage}):} 
    We deform the shape represented by the Gaussian splats via Laplacian-based techniques, treating the means of Gaussian kernels as a point cloud. 
    By combining a point cloud Laplacian with our neighborhood estimation, we achieve plausible deformations in a proxy-free manner.
    \item \textbf{Surface-Preserving Kernel Adaptation (Sec.~\ref{method:Covariance-Adaptation_stage}):} 
    To preserve visual fidelity during deformation, we introduce a novel Gaussian kernel adaptation method. 
    In this stage, we adapt each kernel to reflect a local geometric change, preserving the underlying surface manifold.
\end{enumerate}

\subsection{Surface-Aware Splat Graph Construction} \label{method:SA-graph}
A key challenge of local neighborhood estimation is filtering out spurious neighbors: primitives that are close in Euclidean space but distant along the underlying surface. 
As illustrated in Fig.~\ref{fig:main}, existing point-based distance metrics, such as Euclidean and Mahalanobis distance, cannot be reliably applied in challenging conditions.
Instead, to establish connectivity exclusively along the surface, we build a graph by defining connectivity as an intersection of Gaussian kernels. 
The core intuition is that, \emph{only Gaussian kernels that are proximate along the surface manifold will intersect}. 
First, given a set of 2D images, we reconstruct the scene with a surface-aligned Gaussian Splatting. 
Specifically, we utilize SpikingGS~\cite{zhang2024spiking} for the superior surface alignment quality of 2D Gaussian primitives. 
Since our approach is built upon the surface alignment property, our method can be extended to various surface-aligned GS methods, as demonstrated in Fig.~\ref{fig:abl_othergs}.

The reconstructed scene is represented by a set of $N$ Gaussian kernels $\{\mathcal{G}_{i}\}_{i=1}^{N}$. 
Each kernel $\mathcal{G}_i$ is parametrized by position $\mathbf{p}_i$, covariance $\Sigma_i$, and opacity $\alpha_i$. 
The covariance matrix $\Sigma_i$ is factorized into a rotation matrix $\mathbf{R}_i$ and a scale matrix $\mathbf{S}_i$, such that $\Sigma_i=\mathbf{R}_{i}\mathbf{S}_{i}\mathbf{S}_i^{\top}\mathbf{R}_i^{\top}$. 
To identify the intersections between these kernels, we define an occupancy region $\Omega_i$ for each Gaussian kernel $\mathcal{G}_{i}$ as the set of points in $\mathbb{R}^3$ where the kernel is considered active. 
The activity of a kernel at point $\mathbf{x}$ is determined by two constraints. 
The first is the activation cut-off threshold $\bar{V}_{i}^{p}$ from a spiking neuron of SpikingGS~\cite{zhang2024spiking}, a learnable parameter that suppresses low-opacity parts as:
\begin{align} \label{FIF_neuron}
\hat{\mathcal{G}}_i(\mathbf{x}) &=
\begin{cases}
0, & \mathcal{G}_i(\mathbf{x}) < \bar{V}_i^p,\\
e^{-\frac12(\mathbf{x}-\mathbf{p}_i)^\top\Sigma_i^{-1}(\mathbf{x}-\mathbf{p}_i)}, & \text{otherwise}.
\end{cases}
\end{align}
The second constraint is that the rendered intensity, $\alpha_i\mathcal{G}_{i}(\textbf{x})$, must exceed the minimum contribution threshold $c$, typically ${1}/{255}$. 
Combining these, we define the occupancy region $\Omega_i$ in 3D canonical space as:
\begin{equation}
\Omega_i \;=\; 
\Bigl\{\,\mathbf{x} \in \mathbb{R}^3 \Bigm| 
{\mathcal{G}}_{i}(\mathbf{x})
>  
  \max\Bigl\{\,
    \  \bar{V}^p_{i}
    ,\,\frac{c}{\alpha_i}
  \Bigr\}
\Bigr\}.
\end{equation}
Since the covariance $\Sigma_i$ is rank-2, occupancy region $\Omega_i$ is an ellipse. With scale matrix $\mathbf{S}_i=\text{diag}\left(\sigma_1, \sigma_2, 0\right)$ and rotation matrix $\mathbf{R}_i=[\mathbf{r}_1\ \mathbf{r}_2 \ \mathbf{r}_3]$, we can rewrite the occupancy region $\Omega_{i}$ as following:
\begin{equation} \label{ellipse_form}
\Omega_i = 
\bigl\{
\mathbf{p}_i + \sqrt{\lambda_i}\,\rho \,
\bigl(
  \sigma_{1}\cos\theta\,\mathbf{r}_{1}
  \;+\;
  \sigma_{2}\sin\theta\,\mathbf{r}_{2}
\bigr)
\bigm|
\rho \in [0,1)
\bigr\}
\end{equation}
with $\lambda_i = -2 \max\bigl\{\ln|\bar V_{i}^p|,\;\ln(\frac{c}{\alpha_i})\bigr\}$ and $\theta \in (0, 2\pi]$. This elliptic form provides a direct way of estimating the spatial relationships between Gaussian kernels. To determine whether two Gaussian kernels $\mathcal{G}_i$ and $\mathcal{G}_j$ intersect, we compute their normal-wise offset $\delta_{ij}$ from $\mathcal{G}_i$ to $\mathcal{G}_j$:
\begin{equation} \label{normal_wise_offset}
\delta_{ij} \coloneqq \inf\bigl\{\,|t|
\mid \exists\,\mathbf{x}\in\Omega_i:\;\mathbf{x} + t\,\mathbf{r}_{i,3}\in\Omega_j
\bigr\}
\end{equation}
with $\inf\emptyset = +\infty.$ While a zero offset $\delta_{ij}=0$ denotes a perfect intersection, we consider two Gaussians to be intersecting if their offset $\delta_{ij}$ falls below a tolerance $\epsilon$. This tolerance is necessary to account for minor gaps between adjacent Gaussian kernels, which can arise from the joint optimization of photometric and geometric objectives. 

\subsection{Laplacian-Based Deformation}
\label{method:Deformation_stage}
In this section, we detail the Laplacian-based deformation stage. To achieve proxy-free editing, we directly build Laplacian on the collection of Gaussian kernels, treating them as a point cloud. 
This enables the direct application of Laplacian-based deformation techniques to the reconstructed scene.
Specifically, we construct the Laplacian operator on the set of Gaussian kernels' mean $\{\mathbf{p}_{i}\}_{i=1}^{N}$. 
In this process, we integrate our surface‑aware splat graph into off‑the‑shelf point cloud Laplacian methods. 
Unlike existing kNN-based methods that only consider spatial proximity, we leverage the geodesic distance on our splat graph to ensure the Laplacian $\Delta$ reflect underlying surface faithfully.
We use graph geodesic distance rather than raw graph connectivity for robustness against missing edges.
For efficiency, we compute the cotangent Laplacian on a local triangulation using NonManifold~\cite{nonmanifold}. With the resulting Laplacian operator, we deform the means of kernels $\{\mathbf{p}_{i}\}_{i=1}^{N}$ through Laplacian-based deformation map $\phi_{\Delta}$ as:
\begin{equation}
\mathcal{G}_{\phi_{\Delta}}(\mathbf{x})=e^{-\frac12(\mathbf{x}-\phi_{\Delta}(\mathbf{p}_i))^\top\Sigma_i^{-1}(\mathbf{x}-\phi_{\Delta}(\mathbf{p}_i))}.
\end{equation}
Here, $\phi_{\Delta}$ represents the deformation map obtained by solving a Laplacian-based system such as ARAP~\cite{ARAP} or BBW~\cite{BBW} using Laplacian operator $\Delta$.
However, simply displacing the positions of the primitives leads to severe visual artifacts, because the covariances of the Gaussian kernels are not adapted properly. We therefore present a subsequent kernel adaptation stage, as we detail in the next section.

\begin{figure}[t]
  \centering
  \includegraphics[width=\linewidth]{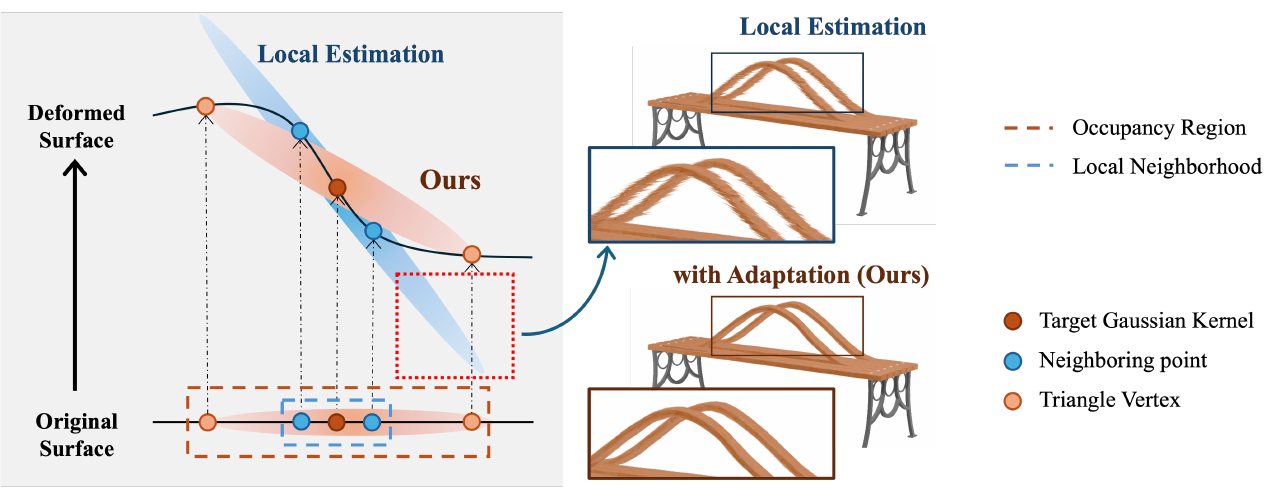}
  \caption{\textbf{Schematic of kernel adaptation methods.} 
  Simply estimating local transformation fails due to a scale mismatch, leading to visual artifacts. Our surface-preserving method avoids this by directly maintaining the original surface coverage.}
  \label{fig:covariance_adaptation}
\end{figure}

\subsection{Surface-Preserving Kernel Adaptation} 
\label{method:Covariance-Adaptation_stage}
Following the deformation stage, we adapt the Gaussian kernels to reflect the local geometric change. However, we observed that a naive approach, such as estimating a local affine transformation from the deformed neighborhood, is insufficient. As illustrated in Fig.~\ref{fig:covariance_adaptation}, when a neighborhood only partially covers a subset of a Gaussian's occupancy region $\Omega$, the local estimation cannot represent the transformation of the entire region. This causes the Gaussian kernel to deviate from the underlying surface, leading to spike-like artifacts. Although prior works have addressed this issue by using geometric proxy such as bounding scales~\cite{MeshGaussian2024}, envelope~\cite{jiang2024vr-gs} or splitting~\cite{huang2024gsdeformerdirectrealtimeextensible}, these are not applicable in our proxy-free scenario. To resolve this, we propose an novel adaptation method that aims to preserve the underlying surface manifold. Our key idea is based on the foundation of 2D Gaussian Splatting~\cite{Huang2DGS2024}, which encourages each kernel to act as a local patch that covers the underlying surface. Instead of estimating a transformation, our goal is to directly adapt each deformed Gaussian kernel $\mathcal{G}_{\phi_{\Delta}, i}$ to preserve its original surface coverage on the newly deformed surface by tracking its occupancy region. However, tracking the entire boundary is computationally intractable. We therefore employ a simpler but representative proxy: the inscribed triangle of maximum area $\mathbf{T}_i=\{\mathbf{t}_{i}^{k}\}_{k=1}^{3}$, defined as:

\begin{equation}
\mathbf t_i^{1}= \mathbf p_i + \mathbf v_{i,1}, \quad \mathbf t_i^{2,3} = \mathbf p_i - \tfrac{1}{2}\,\mathbf v_{i,1} \pm \tfrac{\sqrt{3}}{2}\,\mathbf v_{i,2}
\end{equation}
where $\mathbf{p}_{i}$ is the mean of Gaussian kernel $\mathcal{G}_i$ and $\mathbf{v}_{i,1}$, $\mathbf{v}_{i,2}$ represent the major and minor axis of $\Omega_i$ in Eqn.~\ref{ellipse_form}. The adapted Gaussian kernel and corresponding parameters can then be uniquely recovered from triangle's vertices by computing Steiner circumellipse~\cite{Silvester_2017}. 
However, as the deformation map $\phi_\Delta$ is defined only on the means of the Gaussian kernels $\{\mathbf{p}_i\}_{i=1}^N$, we transfer the displacement $\phi_{\Delta}(\mathbf{p}_{i}) - \mathbf{p}_{i}$ to each triangle vertex $\mathbf{t}_i^k$. Since the displacement on the surface is continuous, we compute the displacement of each vertex $\mathbf{t}_i^k$ using the inverse distance weighting~\cite{shepard1968two} from a set of the adjacent Gaussians, indexed by $\mathcal{M}(\mathbf{t}_i^k)$. The displaced triangle vertex $\mathbf{t}{'}_{i}^{k}$ is then computed as follows:

\definecolor{Gray}{gray}{0.92}
\definecolor{White}{gray}{1.0}
\definecolor{LightLine}{gray}{0.7}
\arrayrulecolor{black} %

\newcolumntype{A}{>{\columncolor{Gray}\footnotesize\centering\arraybackslash}c}
\newcolumntype{B}{>{\columncolor{White}\footnotesize\centering\arraybackslash}c}

\begin{table*}[ht]
    \centering
    \small
    \setlength{\tabcolsep}{4pt} 
    \resizebox{\linewidth}{!}{
    \begin{tabular}{@{}
                    c           
                    !{\color{lightgray}\vrule}
                    A B A B A B A B A B   
                    !{\color{lightgray}\vrule}
                    A B A B A B           
                    !{\color{lightgray}\vrule}
                    A B A B               
                    !{\color{lightgray}\vrule}
                    A B                   
                    !{\color{Black}\vrule width 0.7pt}
                    A B                   
                    @{}}
        \toprule
        
        \textbf{Method}
        & \multicolumn{2}{c!{\color{LightLine}\vrule}}{\footnotesize{Bench}} & \multicolumn{2}{c!{\color{LightLine}\vrule}}{\footnotesize{Bed}} & \multicolumn{2}{c!{\color{LightLine}\vrule}}{\footnotesize{Headphone}} & \multicolumn{2}{c!{\color{LightLine}\vrule}}{\footnotesize{Basket}} & \multicolumn{2}{c!{\color{LightLine}\vrule}}{\footnotesize{Coil}}
        & \multicolumn{2}{c!{\color{LightLine}\vrule}}{\footnotesize{Birdcage}} & \multicolumn{2}{c!{\color{LightLine}\vrule}}{\footnotesize{Windmill}} & \multicolumn{2}{c!{\color{LightLine}\vrule}}{\footnotesize{Ferris Wheel}}
        & \multicolumn{2}{c!{\color{LightLine}\vrule}}{\footnotesize{Harp}} 
        & \multicolumn{2}{c!{\color{LightLine}\vrule}}{\footnotesize{Fence}}
        & \multicolumn{2}{c!{\color{Black}\vrule width 0.7pt}}{Average}
        & \multicolumn{2}{c}{\footnotesize{NS \cite{mildenhall2020nerf}}} \\ 
        \midrule
        
        \multicolumn{25}{c}{\textbf{Proxy-Based Methods}} \\
        \cmidrule(l){1-25} 
        \quad Mani‐GS \cite{gao2024mani}
        & 0.961 & 0.990 
        & 0.808 & 0.927 
        & 0.806 & 0.850 
        & 0.703 & 0.898 
        & 0.605 & 0.816
        & 0.733 & 0.850 
        & 0.822 & 0.891 
        & 0.842 & 0.920
        & 0.785 & 0.817 
        & 0.934 & 0.948 
        & 0.800 & 0.891 
        & 0.801 & 0.825
        \\
        \quad GaussianMesh \cite{MeshGaussian2024}
        & 0.962 & 0.988  
        & 0.811 & 0.921  
        & 0.800 & 0.857  
        & 0.705 & 0.882  
        & 0.598 & 0.821 
        & 0.730 & 0.856 
        & 0.813 & 0.904  
        & 0.829 & 0.932 
        & 0.741 & 0.871  
        & 0.935 & 0.948 
        & 0.792 & 0.898  
        & 0.799 & 0.816
        \\
        \quad SuGaR \cite{guedon2023sugar}
        & 0.862 & 0.650  
        & 0.870 & 0.869 
        & 0.898 & 0.710  
        & 0.935 & 0.878  
        & 0.975 & 0.831 
        & 0.954 & 0.789  
        & 0.903 & 0.873  
        & 0.871 & 0.692 
        & 0.807 & 0.642  
        & 0.858 & 0.640 
        & 0.893 & 0.758  
        & 0.882 & 0.863
        \\
        \midrule

        \multicolumn{25}{c}{\textbf{Proxy-Free Methods}} \\ \cmidrule(l){1-25}
        \quad kNN\,\scriptsize({$k$=10}) \footnotesize\cite{nonmanifold}
        & 0.647  & 0.789 
        & 0.980 & 0.978 
        & 0.952 & 0.981 
        & \textbf{0.993}  & 0.969 
        & 0.895 & 0.930 
        & 0.969  & 0.975 
        & 0.974  & 0.973 
        & 0.963 & 0.973 
        & 0.919  & 0.954 
        & 0.979 & 0.960 
        & 0.927 & 0.948 
        & 0.915 & 0.910
        \\ 
        \quad kNN$^{\dag}$\,\scriptsize({$k$=30}) \footnotesize\cite{nonmanifold}
        & 0.599 & 0.753 
        & 0.946 & 0.943 
        & 0.951 & 0.978 
        & 0.982 & 0.960 
        & 0.779 & 0.808
        & 0.964 & 0.955 
        & 0.955 & 0.977 
        & 0.951 & 0.953
        & 0.718 & 0.813 
        & 0.915 & 0.905
        & 0.876 & 0.904 
        & 0.913 & 0.908
        \\
        \quad{Mahalanobis \scriptsize({$k$=10})}
        \footnotesize\cite{zhou2025laplacebeltramioperatorgaussiansplatting}
        & 0.640 & 0.836 
        & 0.981 & \textbf{0.993} 
        & 0.761 & 0.928 
        & 0.977 & 0.977 
        & 0.866 & 0.927 
        & 0.938 & 0.984 
        & 0.912 & 0.975 
        & 0.943 & 0.975 
        & 0.940 & 0.969 
        & 0.972 & 0.987 
        & 0.893 & 0.955 
        & 0.862 & 0.911
        \\
        \quad{Mahalanobis$^{\dag}$\,\scriptsize({$k$=30})}
        \footnotesize\cite{zhou2025laplacebeltramioperatorgaussiansplatting}
        & 0.606 & 0.791 
        & 0.960 & 0.991 
        & 0.751 & 0.939 
        & 0.966 & 0.973 
        & 0.847 & 0.882
        & 0.952 & 0.977 
        & 0.916 & 0.980 
        & 0.942 & 0.964
        & 0.939 & 0.959 
        & 0.961 & 0.965
        & 0.884 & 0.942 
        & 0.875 & 0.917
        \\
        \quad NeLO \cite{pang2024nelo}
         & 0.671 & 0.785 
         & 0.968 & 0.936 
         & 0.876 & 0.879 
         & 0.968 & 0.885 
         & 0.909 & 0.931
         & 0.952 & 0.952 
         & 0.932 & 0.870 
         & 0.942 & 0.921
         & 0.884 & 0.910 
         & 0.950 & 0.939
         & 0.905 & 0.901 
         & 0.874 & 0.858
         \\
        \midrule
        \multicolumn{25}{@{}c}{\textbf{Ablation Result}} \\ \cmidrule(l){1-25}
        \quad{Splat Graph}
        & 0.996 & 0.994 
        & \textbf{0.993}  & 0.986 
        & 0.945  & 0.976 
        & 0.987  & \textbf{0.983} 
        & \textbf{0.993} & 0.980 
        & 0.974  & 0.991 
        & 0.967  & 0.972 
        & \textbf{0.980} & 0.980 
        & 0.903  & 0.956 
        & 0.995 & 0.993 
        & 0.973 & 0.981 
        & 0.929 & 0.922
        \\ 
        \quad\textbf{Ours} (Geodesic distance)
         & \textbf{0.997} & \textbf{0.998}
         & \textbf{0.993} & \textbf{0.993}
         &  \textbf{0.953} & \textbf{0.982}
         & \textbf{0.993} & 0.977
         & 0.983 & \textbf{0.996}
         & \textbf{0.996} & \textbf{0.997}
         & \textbf{0.980} & \textbf{0.987}
         & 0.965 & \textbf{0.983}
         & \textbf{0.982} & \textbf{0.980}
         & \textbf{0.999} & \textbf{1.000}
         & \textbf{0.985} & \textbf{0.988} 
         & \textbf{0.931} & \textbf{0.924}
         \\
        \bottomrule
    \end{tabular}
    } 
    \caption{\textbf{Quantitative comparison of deformation quality.} We report 3DPCK on our benchmark and NeRF-synthetic \cite{mildenhall2020nerf} at a threshold of 0.075 (higher is better). The score for each category is the average performance across all handles defined for objects in that category. Gray columns correspond to ARAP \cite{ARAP}, while white columns show BBW \cite{BBW} Best scores are in bold. $\dag$ indicates default parameter settings.}
    \label{tab:comparison_final}
\end{table*}

\begin{equation}
\mathbf{t}{'}_{i}^{k}=\mathbf{t}_{i}^{k} +\frac{\sum_{j\in\mathcal{M}(\mathbf{t}_{i}^{k})}\bigl(w_{ij}\bigl(\phi_{\Delta}(\mathbf p_j)-\mathbf{p}_j\bigr)\bigr)}{\sum_{j\in\mathcal{M}(\mathbf{t}_{i}^{k})} w_{ij}}
\end{equation}
where $w_{ij}$ is $\lVert \mathbf t_{i}^{k}-\mathbf p_j\rVert_2^{-1}$.
The adjacent primitives $\mathbf{p}_j \in \mathcal{M}(\mathbf{t}_i^k)$ are chosen based on graph distance on our splat graph to ensure adjacency of the surface.
This adaptation is crucial to ensure that the deformed Gaussian kernels accurately cover their corresponding coverage on the deformed surface, preserving geometric structure and visual fidelity as shown in Fig.~\ref{fig:abl_adaptation} and Fig.~\ref{fig:suppl_adap_more_result}.

\section{Experiments}
We evaluate our framework by applying two representative Laplacian-based deformation techniques, As-Rigid-As-Possible (ARAP)~\cite{ARAP} and Bounded Biharmonic Weights (BBW)~\cite{BBW}, to diverse objects with intricate geometries. We provide the implementation details and further analysis in the supplementary materials.

\subsection{Experiment Setup.} \label{experiment_setup}
\paragraph{Dataset \& Benchmark.} To the best of our knowledge, no suitable dataset exists for evaluating geometric deformations of scenes reconstructed from 2D images. We therefore introduce a new benchmark, specifically curated to assess the fidelity and robustness of deformation. Our benchmark comprises 50 textured synthetic objects across 10 categories, sourced from ShapeNet~\cite{ShapeNet}, Objaverse~\cite{objaverse}, and Sketchfab~\cite{Pinson2011Sketchfab} datasets. 
The number of objects and the breadth of categories in this benchmark are comparable to standard 3D reconstruction benchmarks~\cite{mildenhall2020nerf, DTU}, providing sufficient coverage for evaluating deformation.

For each object, we provide a ground-truth mesh, 100 multi-view rendered images, and a set of interaction handles. 
To test the robustness efficiently, the handles are annotated in geometrically ambiguous regions manually. An average of 14.8 handles are provided per object, for a total of 740 deformation handles across the benchmark. 
In addition, we also conduct a evaluation on NeRF-Synthetic dataset~\cite{mildenhall2020nerf}. The handles for NeRF-Synthetic dataset are sampled by farthest point sampling. Further details of the benchmark are available in the supplementary materials.
\vspace{-2ex} 
\paragraph{Baselines.} We evaluate our framework against two distinct categories of approaches: proxy-based methods and proxy-free methods. For the first category, we compare with the leading methods, Mani-GS~\cite{gao2024mani}, GaussianMesh~\cite{MeshGaussian2024}, and SuGaR~\cite{guedon2023sugar}. For our proxy-free baselines, we compare against existing point-based Laplacian methods including standard kNN-based NonManifold~\cite{nonmanifold}, LBO-GS~\cite{zhou2025laplacebeltramioperatorgaussiansplatting} and NeLO~\cite{pang2024nelo}.
For a fair comparison, we use the optimal hyperparameter for the number of neighbors $k$ in kNN search. We set $k=10$, which we found to be a robust value that avoids graph fragmentation while maintaining locality.
Additionally, our kernel adaptation stage (Sec.~\ref{method:Covariance-Adaptation_stage}) is applied to all proxy-free methods for the qualitative evaluation.

\vspace{-2ex} 
\begin{figure*}[!t]
    \centering
    \setlength{\tabcolsep}{3pt}
    \resizebox{\linewidth}{!}{%
    \begin{tabular}{@{}
        >{\centering\arraybackslash}m{0.02\linewidth}
        >{\centering\arraybackslash}m{0.03\linewidth}
        >{\centering\arraybackslash}m{0.18\linewidth}|
        >{\centering\arraybackslash}m{0.18\linewidth}
        >{\centering\arraybackslash}m{0.18\linewidth}
        >{\centering\arraybackslash}m{0.18\linewidth}|
        >{\centering\arraybackslash}m{0.18\linewidth}
        >{\centering\arraybackslash}m{0.18\linewidth}
        >{\centering\arraybackslash}m{0.18\linewidth}
        >{\centering\arraybackslash}m{0.18\linewidth}
    @{}}
    
    & & 
    \multicolumn{1}{>{\centering\arraybackslash}m{0.18\linewidth}}{\multirow{2}{*}{\large\textbf{Source}}} & 
    \multicolumn{3}{c}{\textbf{Proxy-Based}} & 
    \multicolumn{3}{c}{\textbf{Proxy-Free}} \\
    \cmidrule(lr){4-6} \cmidrule(lr){7-9}
    
    & & & 
    \normalsize\textbf{Mani-GS \cite{gao2024mani}} & 
    \normalsize\textbf{GaussianMesh \cite{MeshGaussian2024}} & 
    \normalsize\textbf{SuGaR \cite{guedon2023sugar}} & 
    \normalsize\textbf{kNN \cite{nonmanifold}} & 
    \normalsize\textbf{NeLO \cite{pang2024nelo}} & 
    \normalsize\textbf{Ours} \\
    \toprule

    \multirow{3}{*}{\large\adjustbox{center, right=-6.4cm, angle=90, valign=m}{\textsc{As-Rigid-As-Possible \cite{ARAP}}}}
    & \adjustbox{center, right=0.25cm, angle=90, valign=m}{\textbf{Birdcage}}
    & \includegraphics[width=\linewidth, valign=m]{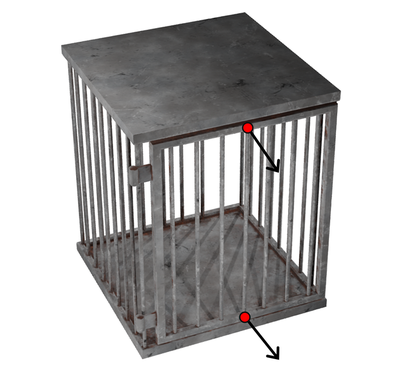}
    & \includegraphics[width=\linewidth, valign=m]{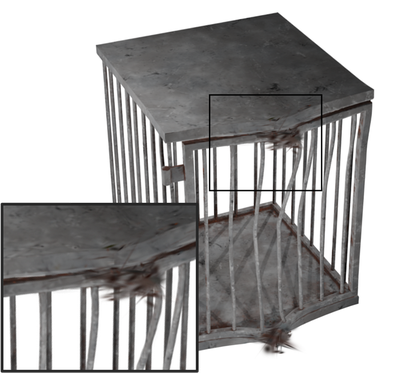}
    & \includegraphics[width=\linewidth, valign=m]{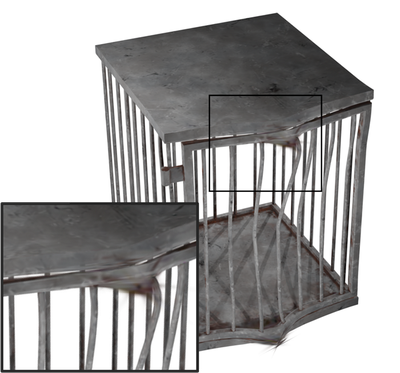}
    & \includegraphics[width=\linewidth, valign=m]{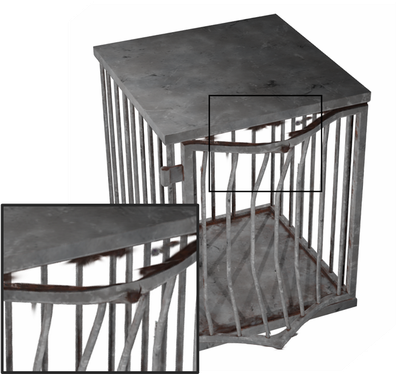}
    & \includegraphics[width=\linewidth, valign=m]{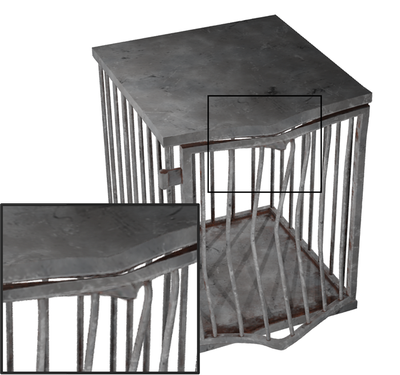}
    & \includegraphics[width=\linewidth, valign=m]{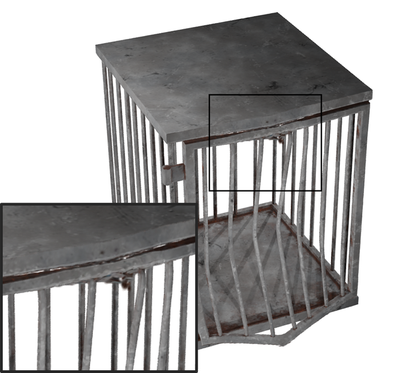}
    & \includegraphics[width=\linewidth, valign=m]{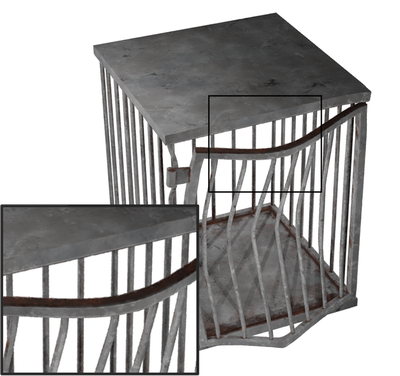} \\ [3pt]
    
    & \adjustbox{center, right=0.3cm, angle=90, valign=m}{\textbf{Harp}}
    & \includegraphics[width=\linewidth, valign=m]{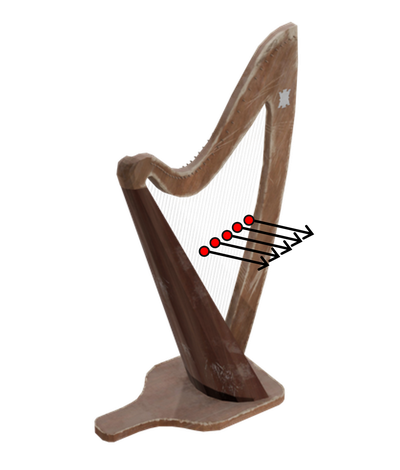}
    & \includegraphics[width=\linewidth, valign=m]{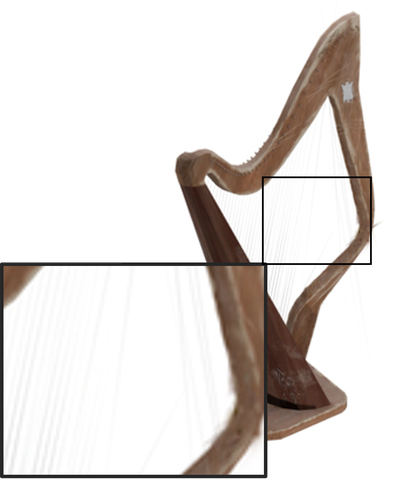}
    & \includegraphics[width=\linewidth, valign=m]{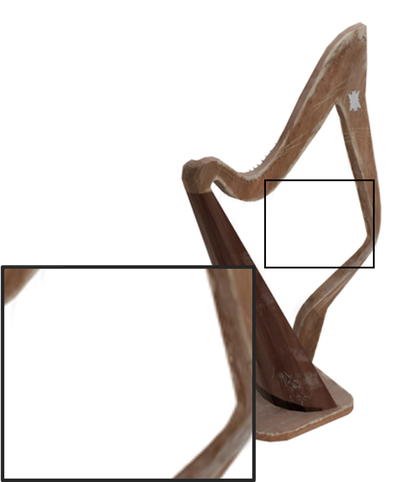}
    & \includegraphics[width=\linewidth, valign=m]{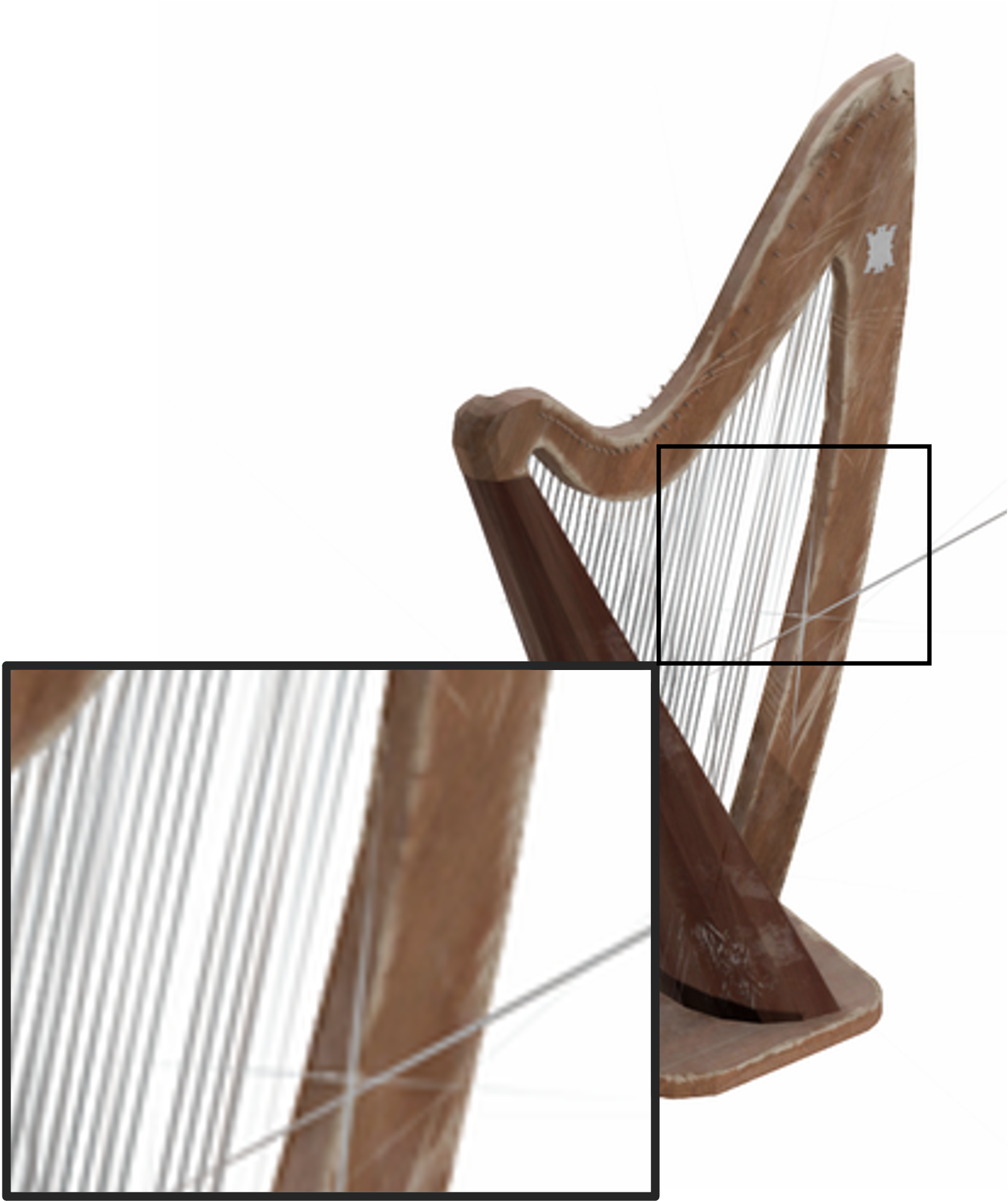}
    & \includegraphics[width=\linewidth, valign=m]{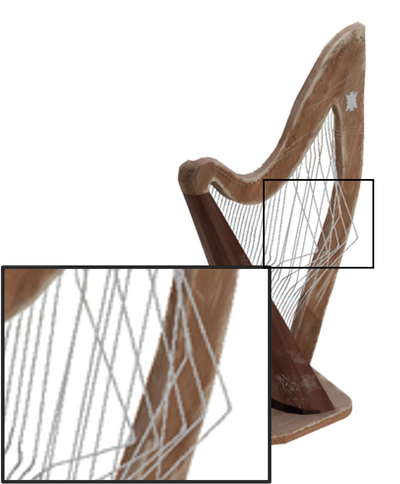}
    & \includegraphics[width=\linewidth, valign=m]{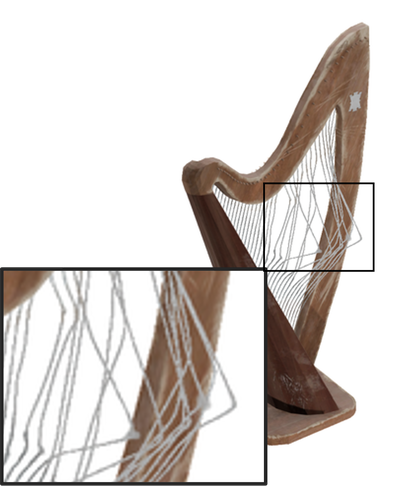}
    & \includegraphics[width=\linewidth, valign=m]{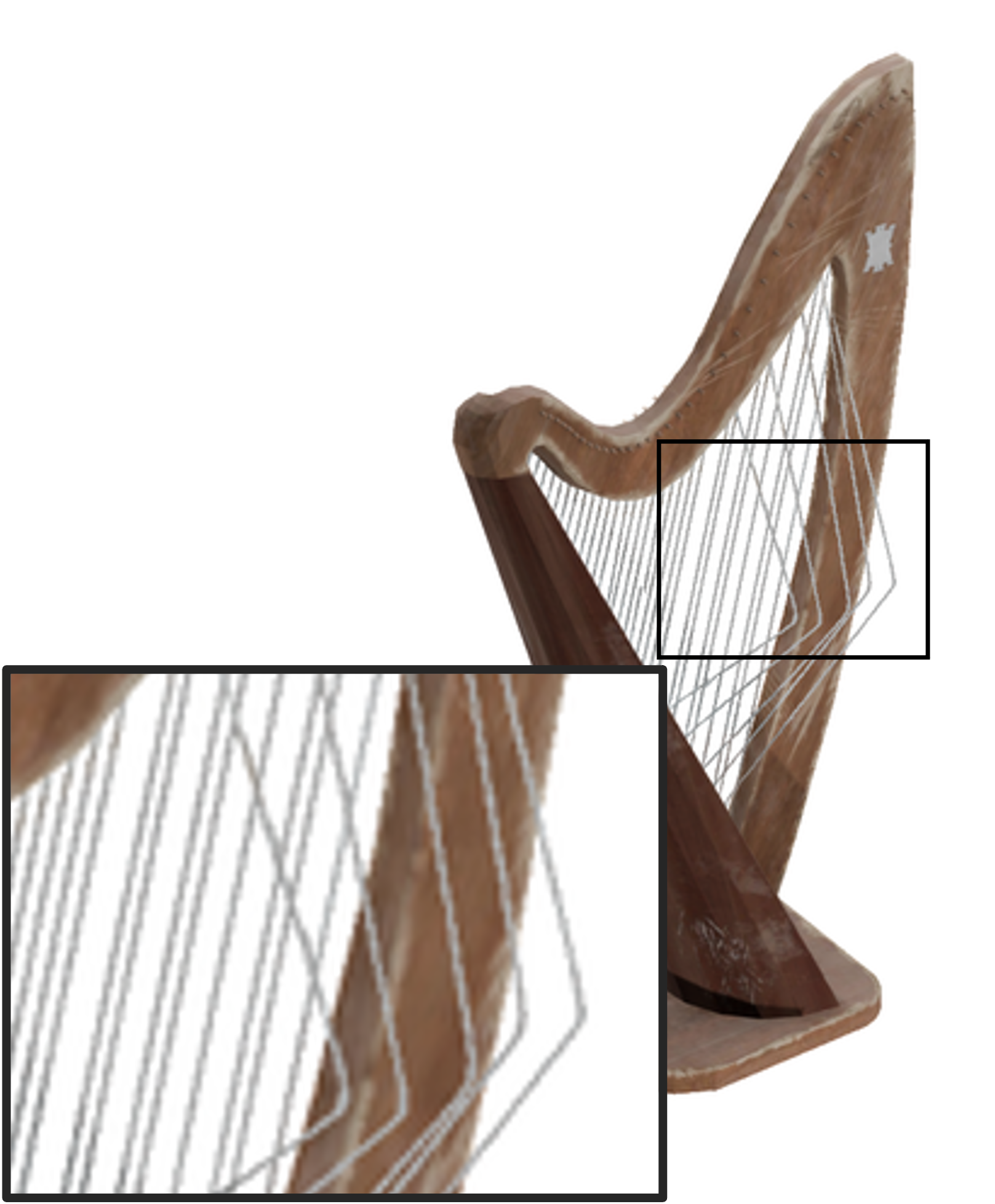} \\ [3pt]
    
    & \adjustbox{center, right=0.3cm, angle=90, valign=m}{\textbf{Windmill}}
    & \includegraphics[width=\linewidth, valign=m]{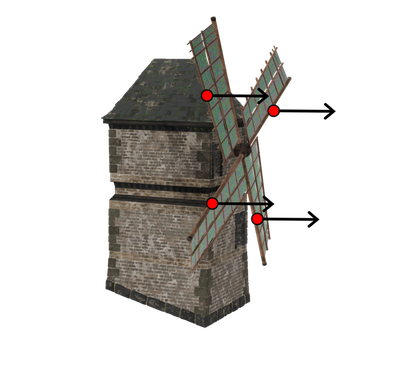}
    & \includegraphics[width=\linewidth, valign=m]{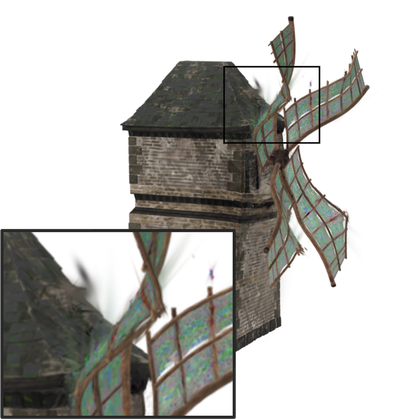}
    & \includegraphics[width=\linewidth, valign=m]{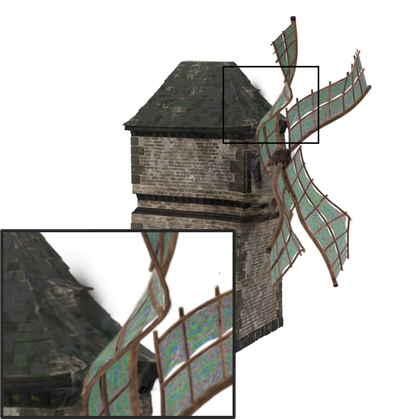}
    & \includegraphics[width=\linewidth, valign=m]{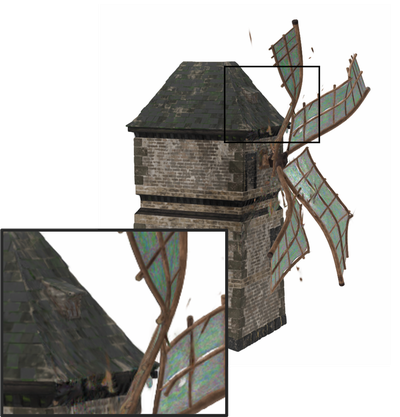}
    & \includegraphics[width=\linewidth, valign=m]{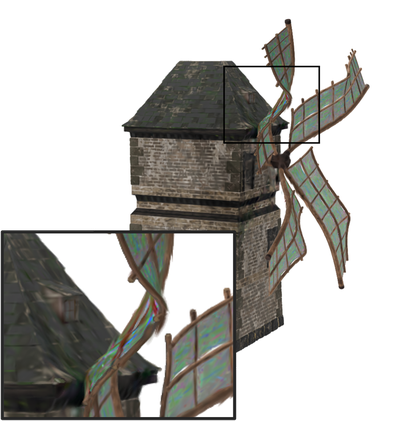}
    & \includegraphics[width=\linewidth, valign=m]{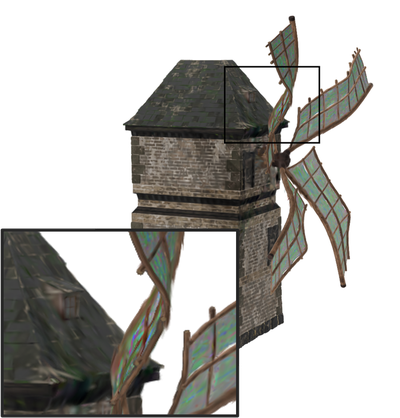}
    & \includegraphics[width=\linewidth, valign=m]{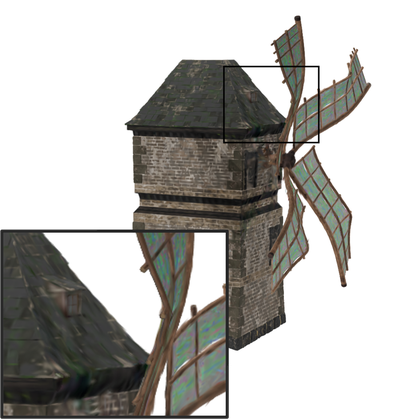} \\

    & \adjustbox{center, right=0.3cm, angle=90, valign=m}
    {\textbf{Bench 1}}
    & \includegraphics[width=\linewidth, valign=m]{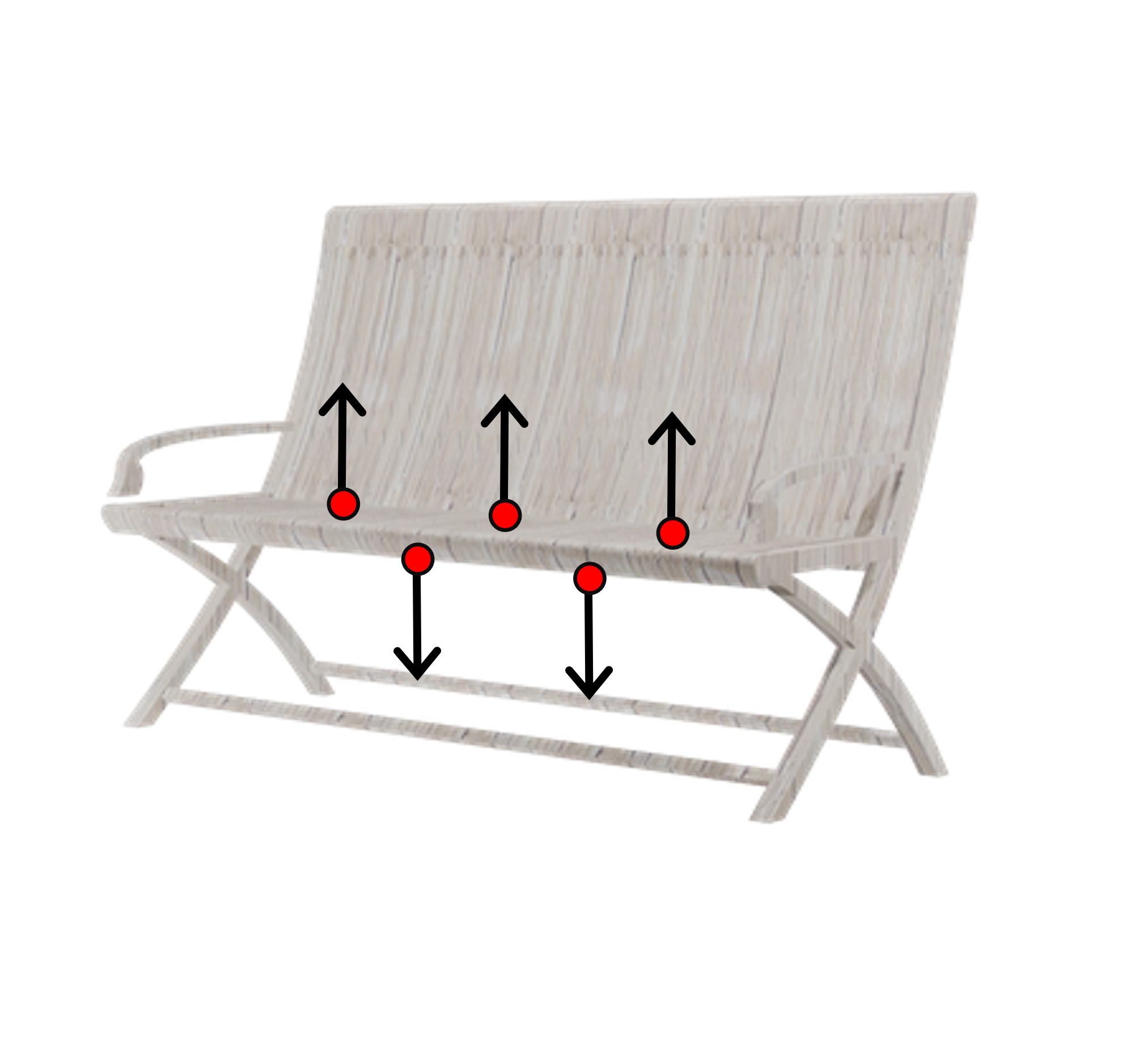}
    & \includegraphics[width=\linewidth, valign=m]{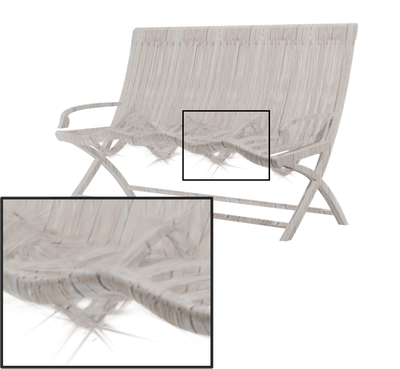}
    & \includegraphics[width=\linewidth, valign=m]{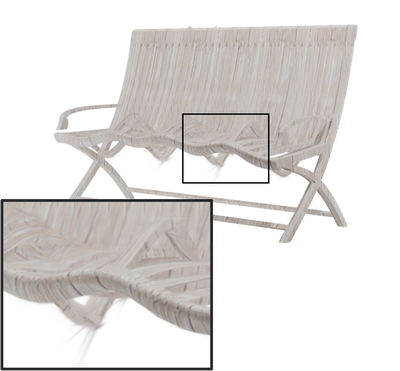}
    & \includegraphics[width=\linewidth, valign=m]{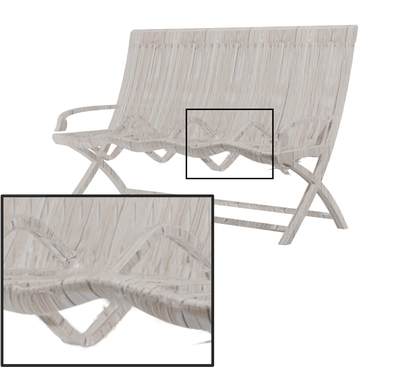}
    & \includegraphics[width=\linewidth, valign=m]{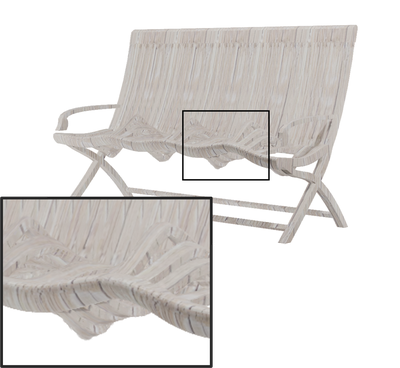}
    & \includegraphics[width=\linewidth, valign=m]{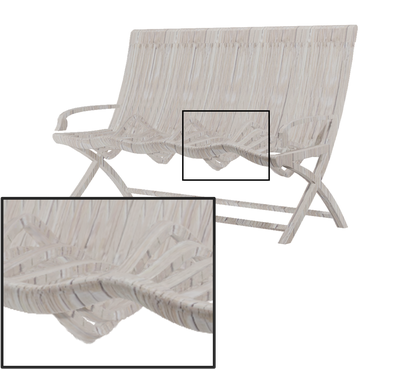}
    & \includegraphics[width=\linewidth, valign=m]{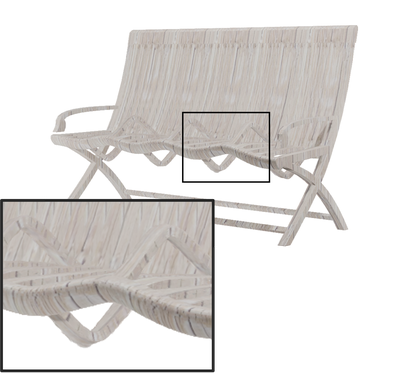} 
    \\ [5pt]

    & \adjustbox{center, right=0.25cm, angle=90, valign=m}
    {\textbf{Bed}}
    & \includegraphics[width=\linewidth, valign=m]{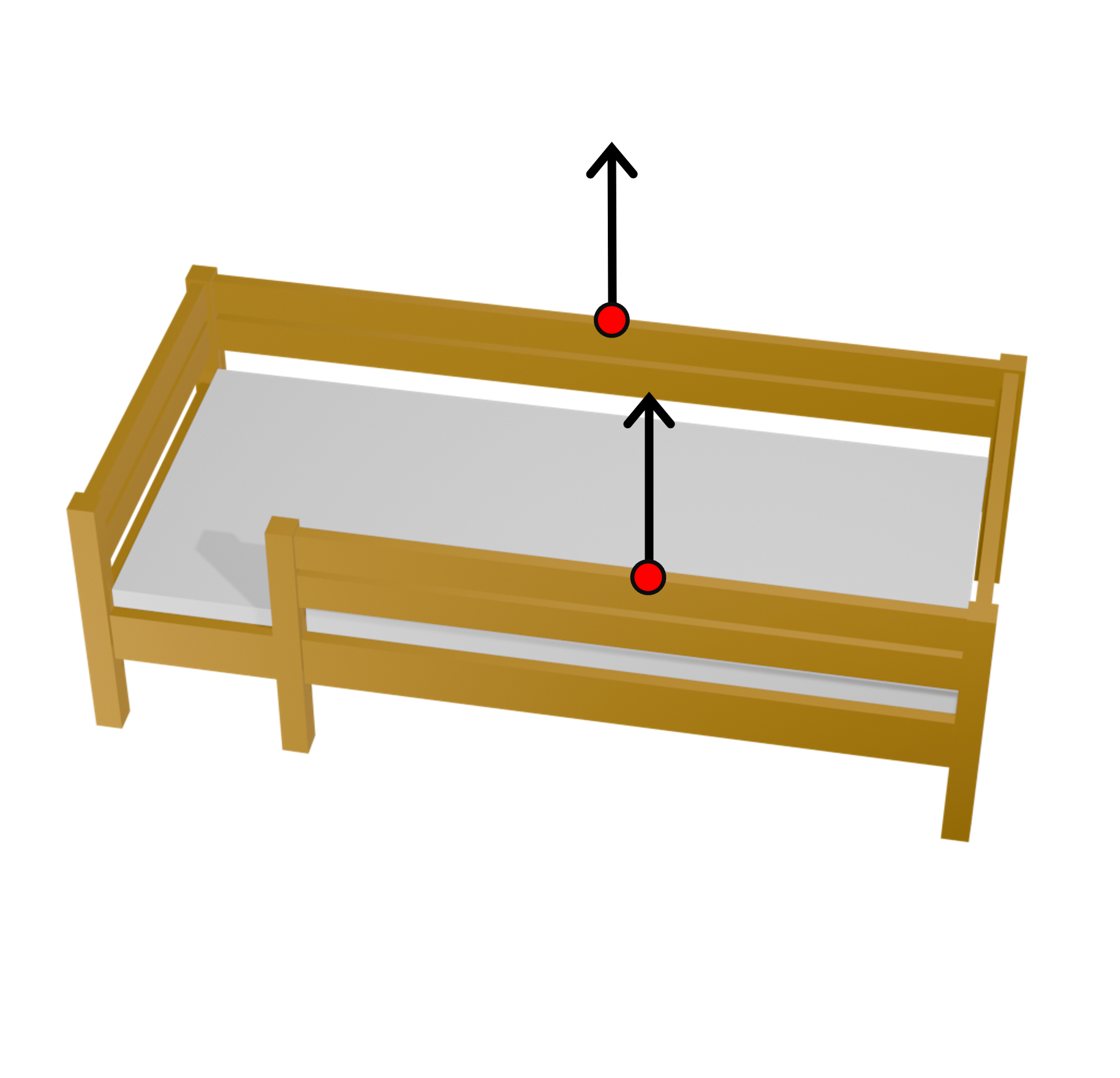}
    & \includegraphics[width=\linewidth, valign=m]{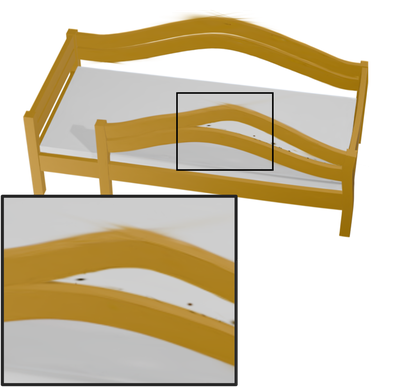}
    & \includegraphics[width=\linewidth, valign=m]{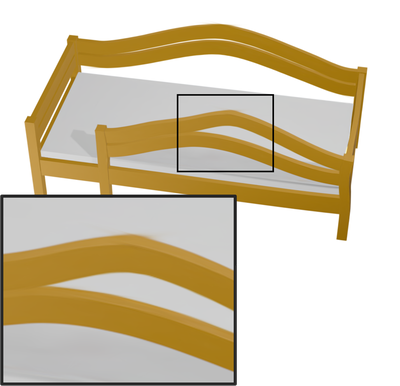}
    & \includegraphics[width=\linewidth, valign=m]{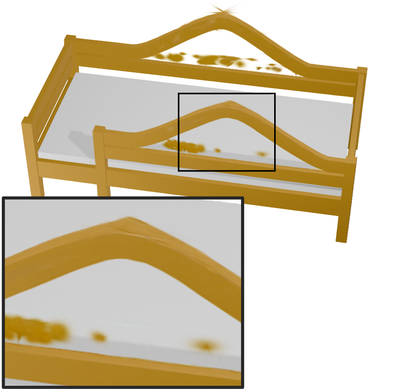}
    & \includegraphics[width=\linewidth, valign=m]{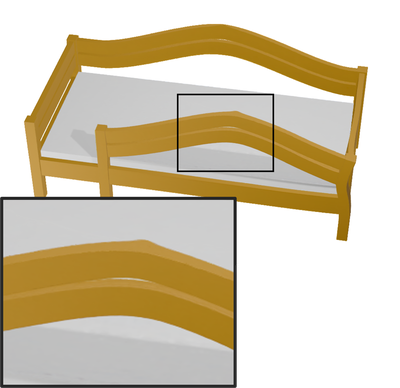}
    & \includegraphics[width=\linewidth, valign=m]{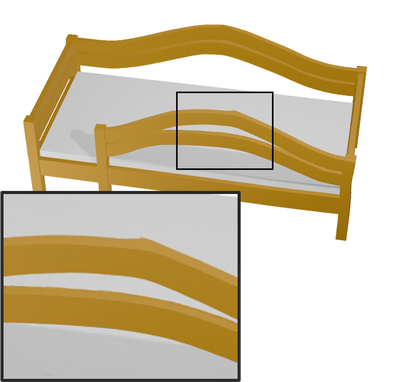}
    & \includegraphics[width=\linewidth, valign=m]{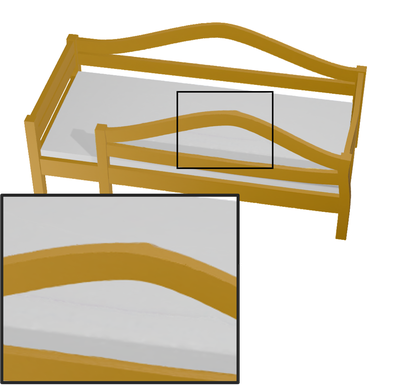} 
    \\ [7pt]
    \midrule
    \multirow{2}{=}{\large\adjustbox{center, right=-3.2cm, angle=90, valign=m}{\textsc{Bounded Biharmonic Weights \cite{BBW}}}}

    & \adjustbox{center, right=0.3cm, angle=90, valign=m}{\textbf{Coil}}
    & \includegraphics[width=\linewidth, valign=m]{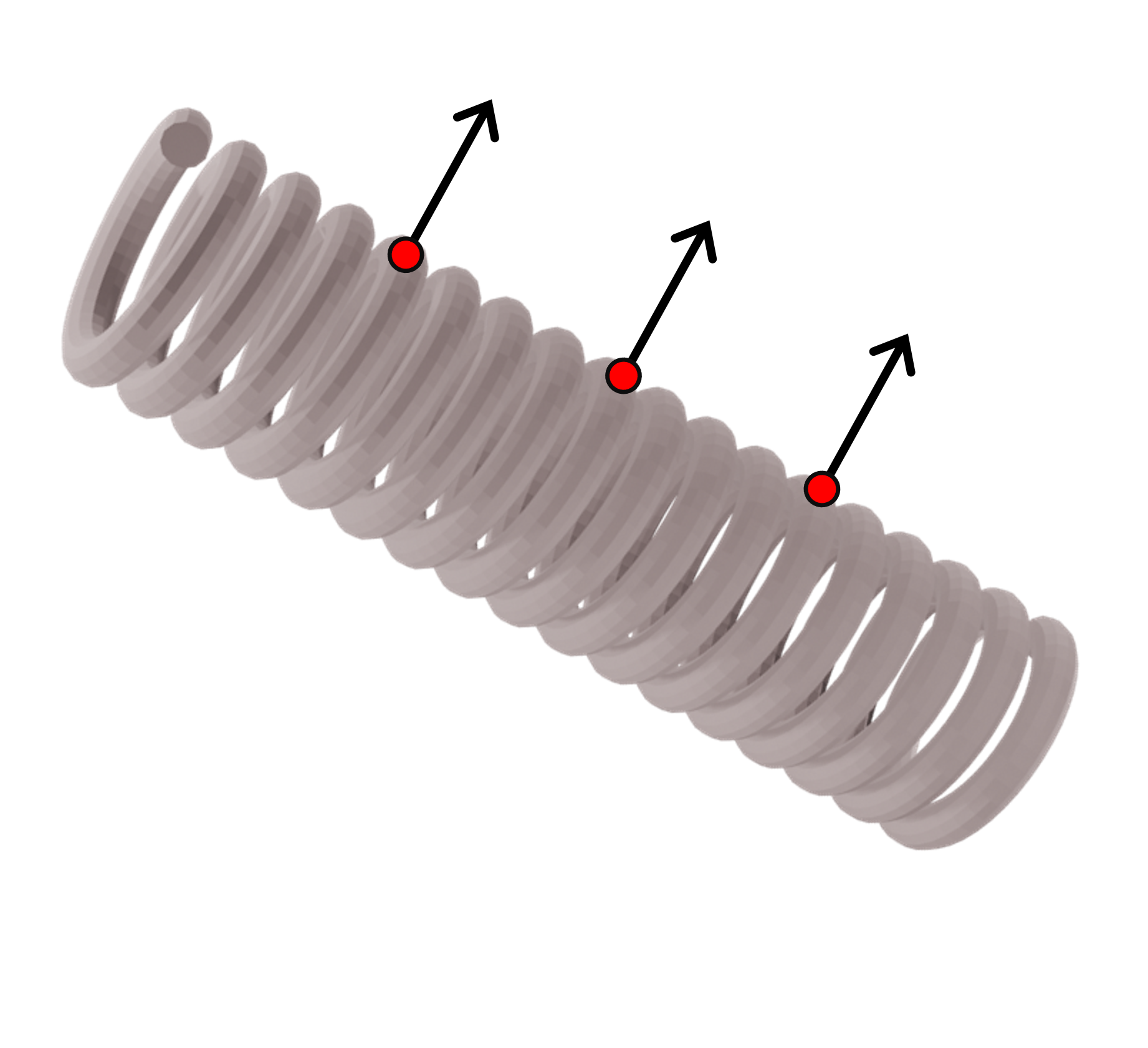}
    & \includegraphics[width=\linewidth, valign=m]{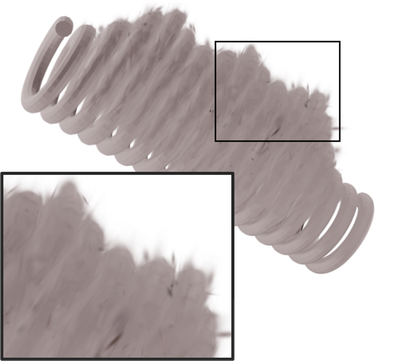}
    & \includegraphics[width=\linewidth, valign=m]{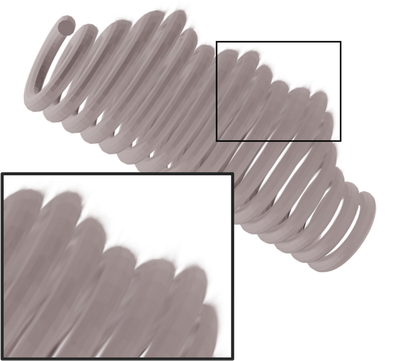}
    & \includegraphics[width=\linewidth, valign=m]{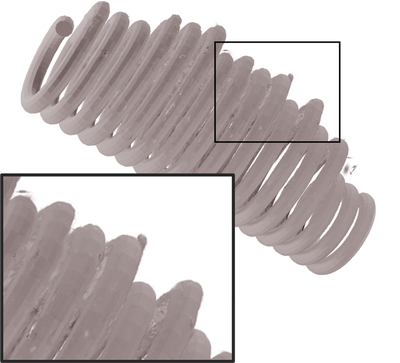}
    & \includegraphics[width=\linewidth, valign=m]{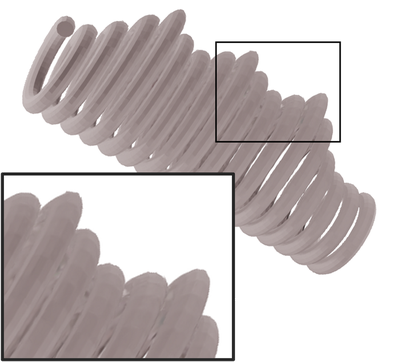}
    & \includegraphics[width=\linewidth, valign=m]{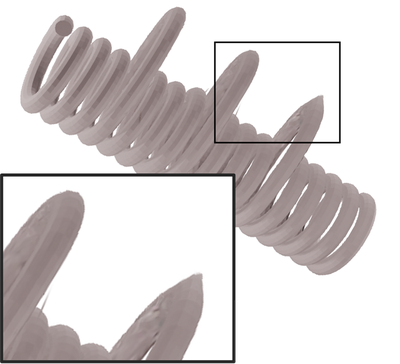}
    & \includegraphics[width=\linewidth, valign=m]{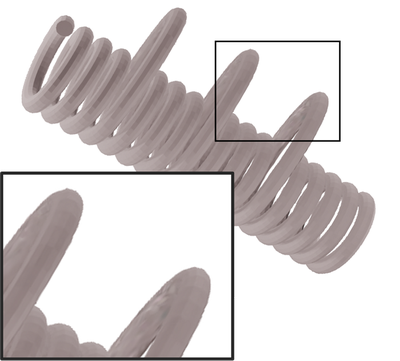} \\ [3pt]
    
    & \adjustbox{center, right=0.25cm, angle=90, valign=m}{\textbf{Bench 2}}
    & \includegraphics[width=\linewidth, valign=m]{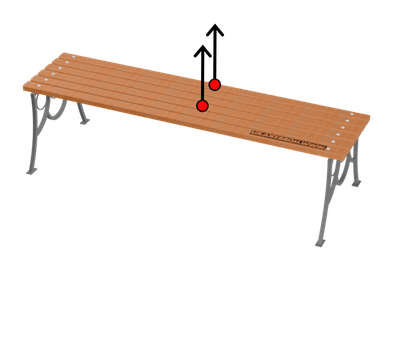}
    & \includegraphics[width=\linewidth, valign=m]{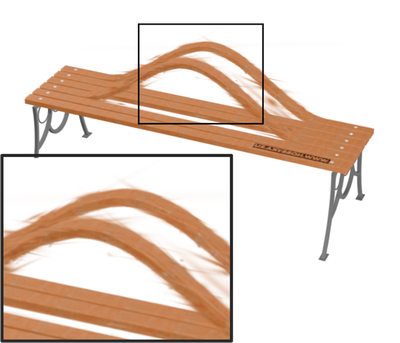}
    & \includegraphics[width=\linewidth, valign=m]{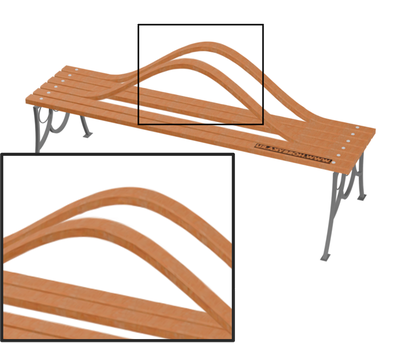}
    & \includegraphics[width=\linewidth, valign=m]{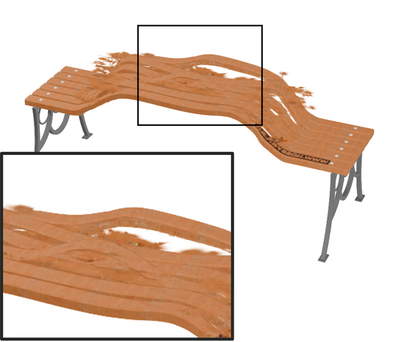}
    & \includegraphics[width=\linewidth, valign=m]{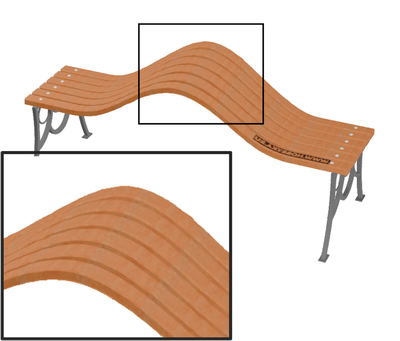}
    & \includegraphics[width=\linewidth, valign=m]{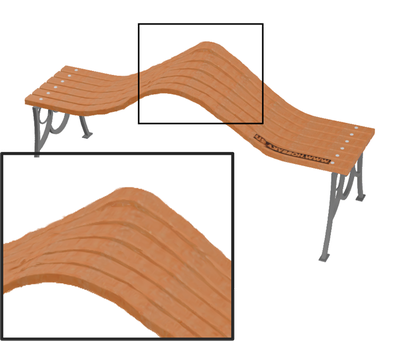}
    & \includegraphics[width=\linewidth, valign=m]{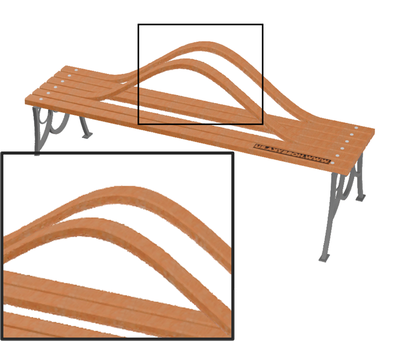} \\ [5pt]
    
    & \adjustbox{center, right=0.25cm, angle=90, valign=m}{\textbf{Fence}}
    & \includegraphics[width=\linewidth, valign=m]{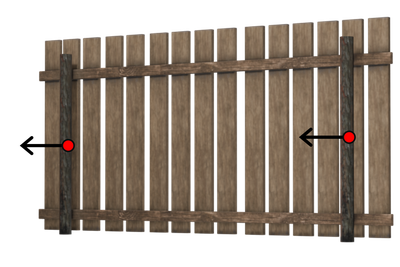}
    & \includegraphics[width=\linewidth, valign=m]{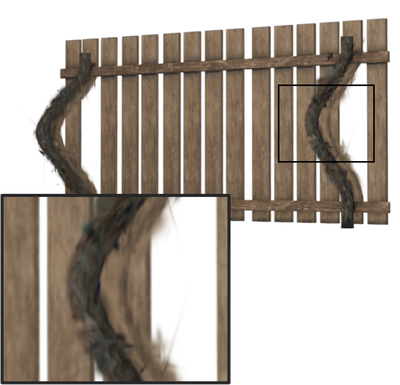}
    & \includegraphics[width=\linewidth, valign=m]{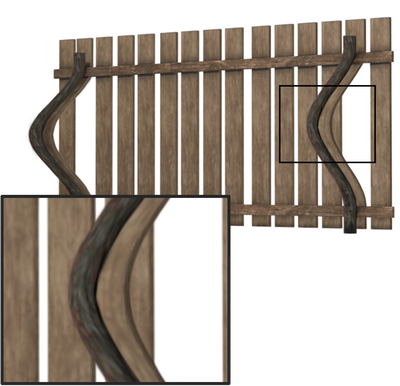}
    & \includegraphics[width=\linewidth, valign=m]{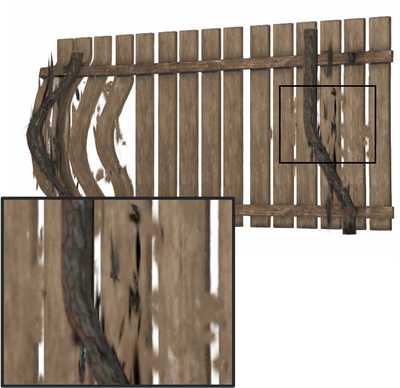}
    & \includegraphics[width=\linewidth, valign=m]{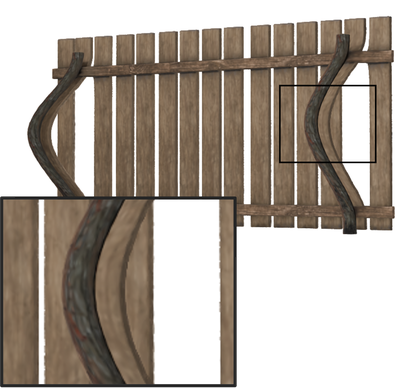}
    & \includegraphics[width=\linewidth, valign=m]{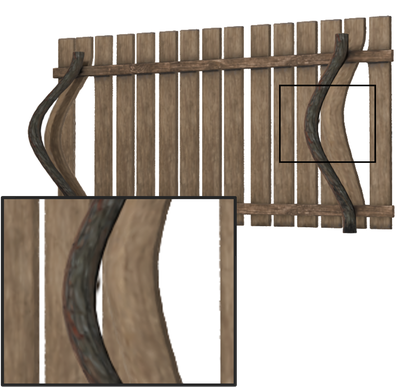}
    & \includegraphics[width=\linewidth, valign=m]{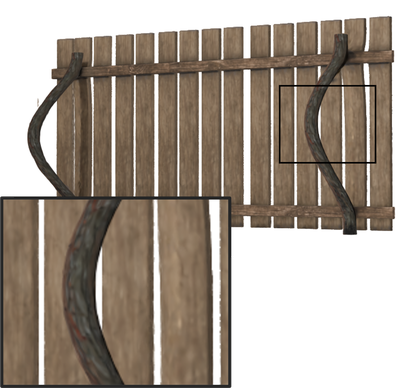} \\
    
    \\
    \bottomrule
    \end{tabular}
    }
    \caption{
    \textbf{Qualitative comparison of deformation results.}
    The red dots denote the interaction handles, with arrows indicating the edit direction.
    The top five rows show results using the ARAP \cite{ARAP}, while the bottom three rows use BBW \cite{BBW}. 
    Our method demonstrates superior visual fidelity and geometric consistency compared to baselines, especially in preserving fine details and preventing artifacts across both deformation techniques.}
    \label{fig:qualitative_results}
\end{figure*}

\paragraph{Evaluation Method.} 
To assess deformation quality, we apply identical transformations to both the ground truth mesh and Gaussian splats. 
For a standardized evaluation, we assign a translation of repulsive direction for each handle using a local PCA of its neighbors, selecting the eigenvector with the smallest eigenvalue.
Unfortunately, mesh processing steps such as watertighting~\cite{manifold_watertight} and remesh~\cite{remesh} required for Laplacian construction discard UV coordinates and textures, As a result, reliable ground truth renderings are no longer available, making standard image-based metrics like PSNR or LPIPS inapplicable.

For this reason, we evaluate the results from two complementary perspectives: 1) topological consistency and 2) visual fidelity. To measure topological consistency in 3D space, we utilize 3D Percentage of Correct Keypoints (3DPCK)~\cite{PCK}, a standard metric in keypoint tracking~\cite{moon2019camera,jiang2020coherent,wang2020hmor,cheng2021graph}. Keypoints are sampled by farthest point sampling around each handle on the ground truth meshes and paired to nearest Gaussian kernels. A deformed keypoint is counted as correct if distance to the mean of paired Gaussian kernel $\mathbf{p}_i$ lies within a threshold. We employ 100 keypoints for computing 3DPCK.
We select 3DPCK instead of metrics for point clouds such as Chamfer Distance for fair comparison, since 3DPCK does not penalize the transparent, but misplaced primitives common in mesh-based methods.
For visual fidelity, we assess the rendered results qualitatively. 
We provide more details about the evaluation and deformation setting in the supplementary materials.

\subsection{Quantitative Results}
Tab.~\ref{tab:comparison_final} summarizes the 3DPCK results of the baselines and our method. Our approach achieves superior performance across all categories, nearly matching the ground-truth. 
Proxy-based approaches~\cite{MeshGaussian2024, gao2024mani, guedon2023sugar} are fundamentally limited by surrogate quality; topological errors in the mesh generation lead to significant degradation in deformation. 
Other proxy-free baselines~\cite{nonmanifold, pang2024nelo, zhou2025laplacebeltramioperatorgaussiansplatting} also show suboptimal performance, particularly on categories with complex topology such as bench and harp, due to their inaccurate neighborhood estimation. 
Our method also shows superior performance on the NeRF-Synthetic dataset~\cite{mildenhall2020nerf} (last column). We provide further result on different PCK thresholds in supplementary materials.

\subsection{Qualitative Results}
We present our qualitative results in Fig.~\ref{fig:qualitative_results}. 
Our method consistently produces high-fidelity deformations, even on objects with intricate geometries.
In contrast, proxy-based methods are commonly limited by the quality of their proxy, failing to produce plausible deformations for shapes with intricate geometries such as the harp shown in the second row of Fig.~\ref{fig:qualitative_results}.
Furthermore, Mani-GS~\cite{gao2024mani}, which lacks regularization on kernel scales, produces severe spike-like artifacts regardless of template quality. 
GaussianMesh~\cite{MeshGaussian2024} cannot compensate for template defects, leading to missing geometry in the reconstruction. 
While other proxy-free baselines circumvent this dependency, their inaccurate local neighborhood estimation causes distortions in unintended regions.

\begin{figure}[t]
  \centering
  \includegraphics[width=\linewidth]{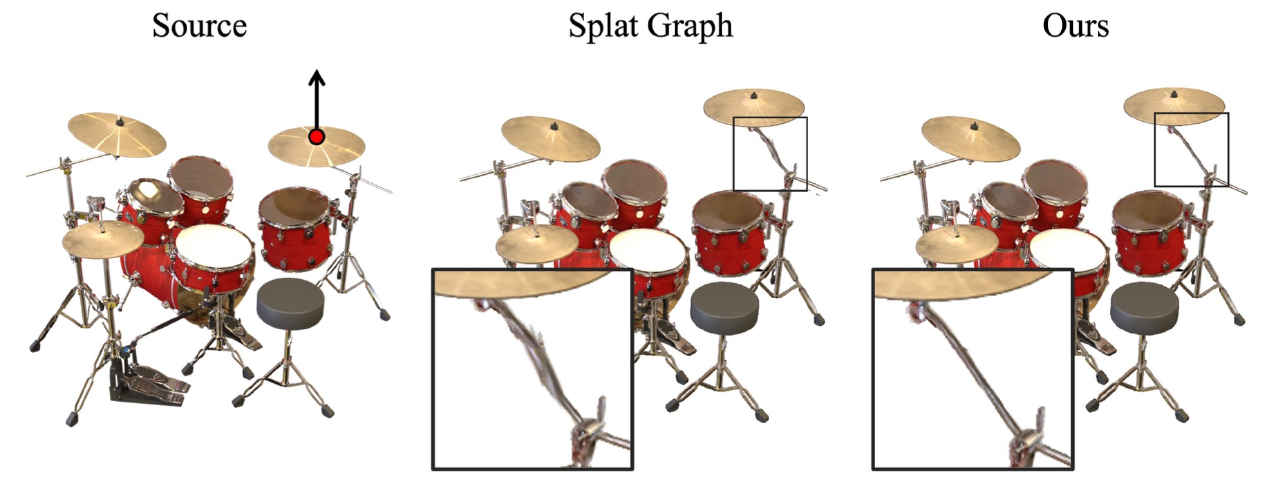}
  \caption{
  \textbf{Comparison of neighborhood estimations.}
  Graph distance based method provides improved robustness than raw graph connectivity on challenging geometries, such as thin structures.}
  \label{fig:abl_neighbor_estimation}
\end{figure}

\begin{figure}[t]
  \centering
  \includegraphics[width=\linewidth]{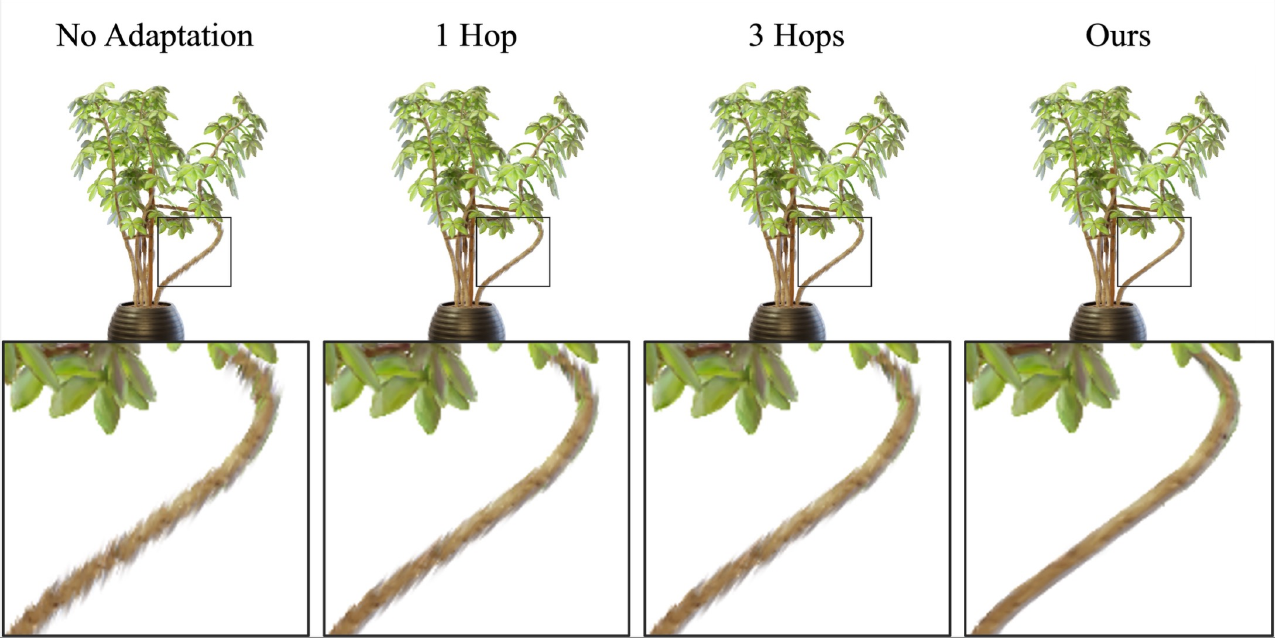}
  \caption{
    \textbf{Comparison of kernel adaptations.}
    $N$ Hop(s) denotes the local transformation estimation based on neighbors within $N$ edges.
    While simple local transformations fail even with wider neighborhoods (3 hops), our method effectively adapts each kernel to maintain high visual quality.
    }
  \label{fig:abl_adaptation}
\end{figure}

\subsection{Ablation Study} \label{exp:ablation}
We perform ablation studies to validate the impact of three key components: graph construction, surface-preserving adaptation, and the choice of the underlying reconstruction method.
\paragraph{Graph Construction.}
Our splat graph-based approach consistently outperforms other point-based methods. However, as illustrated in Fig.~\ref{fig:abl_neighbor_estimation}, a graph defined solely by splat intersection can be sensitive to missing edges, particularly in challenging cases like thin regions. As validated in Tab.~\ref{tab:comparison_final}, using graph distance with neighborhood estimation yields a consistent enhancement over the raw splat graph, showing its improved robustness.

\begin{figure}[t]
  \centering
  \includegraphics[width=\linewidth]{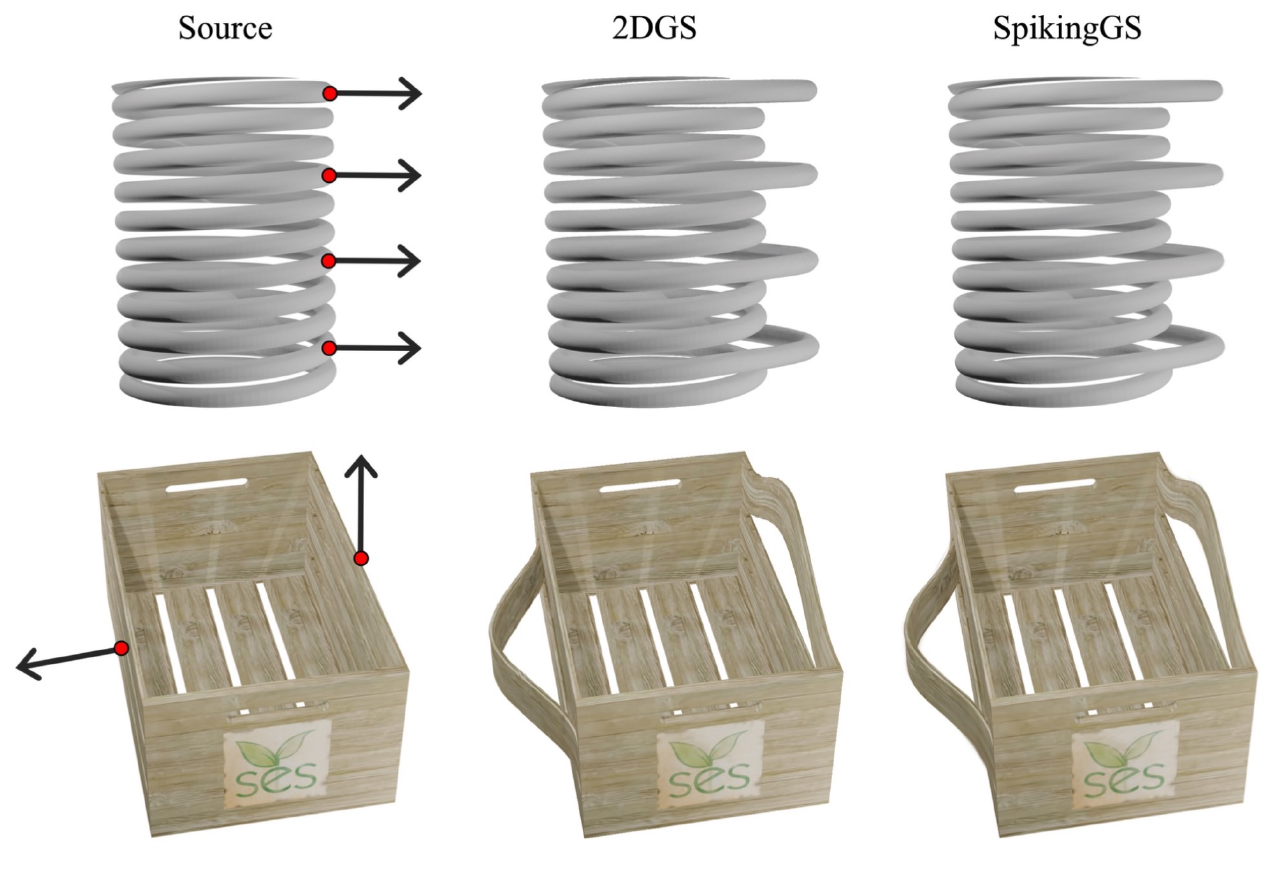}
  \caption{\textbf{Application to other surface-aligned GS methods.} Our framework successfully deforms scenes reconstructed with other surface-aligned GS methods.}
  \label{fig:abl_othergs}
\end{figure}

\vspace{-2ex} 
\paragraph{Surface-Preserving Adaptation.}
We validate the impact of the adaptation stage introduced in Sec.~\ref{method:Covariance-Adaptation_stage}. As shown in Fig.~\ref{fig:abl_adaptation} and Fig.~\ref{fig:suppl_adap_more_result}, noticeable spike artifacts appear since a naive local transformation estimate cannot capture the full extent of elliptical support. Simply expanding the neighborhood (\emph{e.g.,} by increasing the number of hops) does not resolve this fundamental scale mismatch. Our method explicitly re-aligns each kernel with its deformed surface patch, a step that is crucial for preserving visual fidelity.
\vspace{-2ex} 
\paragraph{Generalization to Other Reconstruction Methods.}
To validate the robustness of our framework to the underlying reconstruction method, we apply our deformation framework to scenes generated by 2DGS~\cite{Huang2DGS2024}, a representative 2D primitive-based method. For this experiment, we define the occupancy regions by omitting the spiking threshold $\bar{V}^p$ constraint, which is not present in 2DGS. The qualitative results in Fig.~\ref{fig:abl_othergs} confirm that our approach successfully achieves high-quality geometric deformations, proving its independence from a specific reconstruction method.

\section{Conclusion and Limitations}
In this work, we present SpLap, a proxy-free framework for the high-fidelity geometric deformation of Gaussian splats. Our approach leverages the geometric information inherent in surface-aligned GS, re-interpreting it as a discrete surface representation. With this intuition, we enable the direct application of existing Laplacian-based deformation techniques. Our experiments demonstrate that this approach can achieve a deformation quality that is on par with applying the same edits to the ground truth meshes, thereby circumventing the dependency on geometric proxies. This finding positions Gaussian Splatting not just as a technique for inverse rendering, but as a viable deformable geometric representation in its own right. 

\paragraph{Limitations.} 
Although SpLap achieves high-quality deformation without relying on an external proxy, our approach requires a well-aligned reconstruction. Since our framework does not explicitly refine the reconstruction itself, it may fail to infer intrinsic distances when the underlying Gaussian Splatting (GS) quality is poor. This reliance on the assumption that Gaussian kernels are well-aligned with the surface may limit the applicability of our method to general, unconstrained 3D Gaussian scenes. We leave the extension toward more general and robust deformation handling for future work. Although we restrict our focus to 2D primitive-based surface-aligned GS in this study, our approach is applicable to other geometry-integrated GS methods, provided they offer sufficient reconstruction quality.



{
  \small
  \bibliographystyle{unsrt}   
  \bibliography{references}   
}

\clearpage
\appendix
\setcounter{page}{1}
\maketitlesupplementary
This supplementary material provides additional details, derivations, and experiments to complement the main paper.
We begin by presenting a detailed derivation of the occupancy region in Sec.~\ref{suppl:occ_region_deriv} and details of intersection metric computation between splats in Sec.~\ref{suppl:intersection-check}. We then provide technical details including implementation details and preprocessing in Sec.~\ref{suppl:impl_detail}, followed by specifics of the deformation techniques used in Sec.~\ref{suppl:deformation_detail}. Further details on our proposed benchmark are available in Sec.~\ref{suppl-benchmark-section}. Finally, we present additional experiments and analyses in Sec.~\ref{sec:supl_more_results}, including results under varying 3DPCK thresholds, the effect of the hyperparameter $k$, and further ablations of our surface-preserving adaptation.

\section{Derivation of Occupancy Region} \label{suppl:occ_region_deriv}
We clarify the derivation of the occupancy region $\Omega$ from a Gaussian kernel $\mathcal G$. 
For Gaussian kernel $\mathcal{G}_i$ with mean $\mathbf{p}_{i}$ and covariance matrix 
$\Sigma_i\!=\!\mathbf{R}_{i}\mathbf{S}_{i}\mathbf{S}_i^{\top}\mathbf{R}_i^{\top}$ where rotation matrix $\mathbf{R}_i\!=\![\mathbf{r}_1\ \mathbf{r}_2 \ \mathbf{r}_3]$ and scale matrix $\mathbf{S}_i=\text{diag}\left(\sigma_1, \sigma_2, 0\right)$, 
define
\begin{equation}
T_i:=\max\bigl\{\bar V_i^{p},\,\frac{c}{\alpha_i}\bigr\},
\quad
\lambda_i:= -2\max\bigl\{\ln|\bar V_{i}^p|,\;\ln(\frac{c}{\alpha_i})\bigr\} 
\end{equation}
with opacity $\alpha_i$, adaptive threshold $\bar{V}_i^p$, minimum rendering contribution threshold $c\!=\!1/255$.
Since $\bar V_i^{p}$ is passed through a sigmoid function and entries with $\alpha_i<c$ are discarded; hence $\lambda_i$ is strictly positive.
Then $\mathcal{G}_i(\mathbf{x})>T_i$ is equivalent to
\begin{equation}
(\mathbf{x}-\mathbf{p}_i)^{\!\top}\Sigma_i^{+}(\mathbf{x}-\mathbf{p}_i)<\lambda_i ,
\end{equation}
where $\Sigma_i^{+}$ denotes the Moore–Penrose pseudoinverse of the rank‑2 covariance matrix.  

With $\Sigma_i=\mathbf{R}_i\operatorname{diag}(\sigma_1^{2},\sigma_2^{2},0)\mathbf{R}_i^{\!\top}$ and local
coordinates $\boldsymbol{\xi}:=\mathbf{R}_i^{\!\top}(\mathbf{x}-\mathbf{p}_i)=(\xi_1,\xi_2,0)^{\!\top}$,
we obtain
\begin{equation}
\frac{\xi_1^{2}}{\sigma_1^{2}}+\frac{\xi_2^{2}}{\sigma_2^{2}}<\lambda_i,
\qquad
\xi_3=0 .
\end{equation}
Parametrizing with $\rho\in[0,1)$, $\theta\in(0,2\pi]$ as
$\boldsymbol{\xi}=\sqrt{\lambda_i}\,\rho(\sigma_1\cos\theta,\sigma_2\sin\theta,0)^{\!\top}$
and rotating back yields the following compact expression,
\begin{equation}
\Omega_i=\bigl\{\,
\mathbf{p}_i+ \sqrt{\lambda_i}\,\rho\bigl(
\sigma_1\cos\theta\,\mathbf{r}_1+
\sigma_2\sin\theta\,\mathbf{r}_2
\bigr)
\bigr\}\;
\end{equation}
where $\rho\!\in[0,1)$ and $\theta\!\in(0,2\pi]$.

\section{Details of Intersection Metric Computation}\label{suppl:intersection-check}
In this section, we detail the intersection determination between elliptic occupancy regions
$\Omega$ of Gaussian kernels $\mathcal{G}$ in 3D space.
As in Sec.~\ref{method:SA-graph}, we denote by
$\Omega_i \subset \mathbb{R}^3$ the surface-aligned elliptic occupancy region
associated with $\mathcal{G}_i$, parametrized by its center
$\mathbf{p}_i \in \mathbb{R}^3$, unit normal
$\mathbf{n}_i \in \mathbb{S}^2$, and orthonormal in-plane directions
$\mathbf{r}_{i,1}, \mathbf{r}_{i,2} \in \mathbb{R}^3$ with semi-axis lengths $a_i, b_i > 0$.

For robustness against minor misalignment, we use an
$\epsilon$-intersection criterion based on a normal-wise offset,
instead of a strict binary intersection between splats.
Given a tolerance $\varepsilon > 0$, we say that $\mathcal{G}_i$ and $\mathcal{G}_j$
intersect if $\delta_{ij} \le \epsilon$ where $\delta_{ij}$ measures how far $\Omega_j$ must be translated along
$\mathbf{n}_i$ to touch $\Omega_i$.

\paragraph{Normal-wise offset.}
We consider translations of $\Omega_j$ by $t \mathbf{n}_i$ with $t \in \mathbb{R}$, and define
\begin{equation}
    \delta_{ij}
    :=
    \inf_{t \in \mathbb{R}}
    \bigl\{
        |t|
        \;\big|\;
        \Omega_i \cap (\Omega_j - t \mathbf{n}_i) \neq \emptyset
    \bigr\}.
\end{equation}
By definition, $\delta_{ij} = 0$ if the splats already intersect, and
$\delta_{ij} = +\infty$ if no translation along $\mathbf{n}_i$ can produce a contact.

To evaluate $\delta_{ij}$, we work in a local coordinate system aligned with $\Omega_i$.
For brevity, we denote the in-plane basis of $\mathcal{G}_i$ by $\mathbf{r}_1 := \mathbf{r}_{i,1}$ and $\mathbf{r}_2 := \mathbf{r}_{i,2}$
and keep $\mathbf{n}_i$ as the normal.
We define three scalar coordinate functions
\begin{align}
    X(x) &= \frac{\langle x - \mathbf{p}_i, \mathbf{r}_1 \rangle}{a_i}, \\
    Y(x) &= \frac{\langle x - \mathbf{p}_i, \mathbf{r}_2 \rangle}{b_i}, \\
    Z(x) &= \langle x - \mathbf{p}_i, \mathbf{n}_i \rangle.
\end{align}
In these coordinates, $\Omega_i$ becomes the unit disk in the plane $Z = 0$,
\begin{equation}
    \Omega_i'
    =
    \bigl\{
        (X,Y,Z) \;\big|\;
        X^2 + Y^2 \le 1,\; Z = 0
    \bigr\},
\end{equation}
and $\Omega_j$ is mapped to an elliptic patch $\Omega_j'$ with some projected ellipse
in the $(X,Y)$-plane.
A translation of $\Omega_j$ by $t \mathbf{n}_i$ in 3D space corresponds to a shift of $\Omega_j'$
by $t$ along the $Z$-axis.
Therefore, $\delta_{ij}$ can be written compactly as
\begin{equation}
    \delta_{ij}
    =
    \inf_{x \in \Omega_j}
    \bigl\{
        |Z(x)|
        \;\big|\;
        X(x)^2 + Y(x)^2 \le 1
    \bigr\}.
\end{equation}

\paragraph{Numerical approximation.}
In practice, we approximate the above infimum via sampling.
We uniformly sample points on the boundary of $\Omega_j$ using its ellipse
parameterization (in terms of $\mathbf{r}_{j,1}, \mathbf{r}_{j,2}$), evaluate
$X(x), Y(x), Z(x)$ for each sample, and keep only those satisfying
$X(x)^2 + Y(x)^2 \le 1$, i.e., whose projection lies inside the unit disk.
We additionally sample points on the unit circle $X^2 + Y^2 = 1$,
restrict them to the region covered by the projection of $\Omega_j'$,
and obtain their $Z$-values by interpolation from nearby samples.
Collecting all candidate heights $\{|Z_k|\}$, we use
\begin{equation}
    \hat{\delta}_{ij} = \min_k |Z_k|
\end{equation}
as the numerical estimate of the normal-wise offset, and apply the
$\epsilon$-intersection criterion $\hat{\delta}_{ij} \le \epsilon$
in Sec.~\ref{method:SA-graph}.

\section{Technical Details} \label{suppl:impl_detail}
\label{sec:supl_implementation_detials}
This section describes the detailed implementation of our framework and baselines.

\subsection{Implementation Details} \label{sppl-our-impl-detail}
\paragraph{SpLap.}
Our framework is implemented based on SpikingGS~\cite{zhang2024spiking} using default configuration for NeRF-Synthetic~\cite{mildenhall2020nerf}.

To avoid the intractable quadratic complexity of an all-pairs comparison, we accelerate our intersection queries using a fixed-radius search, which reduces the complexity to approximately linear.
This optimization allows our entire framework to run efficiently on a multi-core CPU alone. For a typical scene of $10^5$ primitives, the one-time preprocessing, such as Laplacian construction, completes in minutes, and each subsequent deformation including kernel adaptation takes less than a minute.

For hyperparameter setting, we use the intersection tolerance $\epsilon=0.005s$ where $s$ is the scale of the bounding box of the scene in our splat graph construction, the number of neighbors $k\!=\!30$ for the Laplacian construction and $k\!=\!3$ for the binding triangle vertex in Gaussian kernel adaptation.
This single hyperparameters configuration is used across all experiments.
By estimating the local neighborhood based on intrinsic distance rather than simple spatial proximity, our method is robust to the choice of the number of neighbors $k$, which is a hyperparameter that is critical for the quality of the graph construction.
We provide an analysis in Sec.~\ref{suppl-effect-hyperparameter}, where we validate this robustness in comparison to a standard kNN baseline.

\paragraph{Baselines.}
For proxy-based baselines, we generate meshes using NeuS2~\cite{neus2} following GaussianMesh~\cite{MeshGaussian2024}. 
To ensure the quality of generated mesh, we set the marching cubes resolution to 1400, which is the maximum supported by GPU memory, and then simplify the generated mesh to have approximately $30k$ vertices on average, following the official implementations. 

We use SpikingGS~\cite{zhang2024spiking} as the reconstruction backbone for all proxy-free baselines. 
For the kNN baseline~\cite{nonmanifold}, we use both the default ($k\!=\!30$) and our optimized ($k\!=\!10$) parameter settings for quantitative results. We show the optimal parameter  ($k\!=\!10$) cases in qualitative results to ensure a fair visual comparison.
For LBO-GS~\cite{zhou2025laplacebeltramioperatorgaussiansplatting}, we apply Mahalanobis distance metric and low-opacity filtering. To ensure numerical stability, the smallest scale component of each kernel is regularized by setting it to one-tenth of the smaller of the other two components.
For NeLO~\cite{pang2024nelo}, we employ the pretrained weights provided by the authors due to the lack of available training data.

\begin{figure}[t!]
  \centering
  \includegraphics[width=\columnwidth]{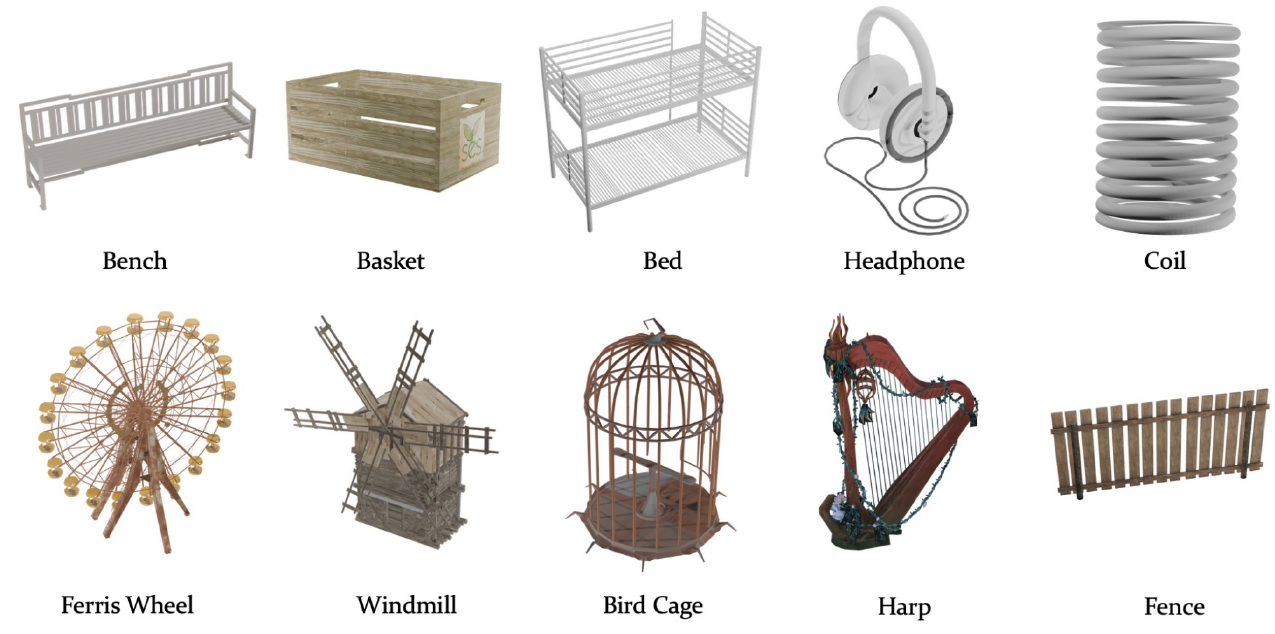}
  \caption{\textbf{Visualization of objects from our benchmark.} Our benchmark includes 50 objects with intricate geometry from 10 categories, sourced from ShapeNet~\cite{ShapeNet}, Objaverse~\cite{objaverse}, and SketchFab~\cite{Pinson2011Sketchfab} to cover a wide range of geometric complexities.}
  \label{fig-suppl-benchmark-vis}
\end{figure}
\begin{table}[t!] 
  \centering
  \vspace{0.3em}
  \resizebox{\linewidth}{!}{%
  \begin{tabular}{lccccc}
    \toprule
    \multirow{2}{*}{Category} & \multirow{2}{*}{\#\,Objects} & \multicolumn{2}{c}{Mesh Resolution} &
    \multirow{2}{*}{\#\,Images} & \multirow{2}{*}{Avg.\ Handles} \\ \cmidrule{3-4}
    & & Vertices & Faces & &  \\ \midrule
    Basket          &  4 & 31$k$ & 63$k$ & 100 & 14.75 \\
    Bench           &  5 & 20$k$ & 40$k$ & 100 & 16.0 \\
    Bed             &  5 & 30$k$ & 59$k$ & 100 & 15.6 \\
    Birdcage        &  5 & 31$k$ & 63$k$ & 100 & 15.8 \\
    Coil            &  6 & 28$k$ & 56$k$ & 100 & 19.5 \\
    Fence           &  4 & 16$k$ &  32$k$ & 100 & 10.5 \\
    Ferris wheel     & 5 & 42$k$ & 87$k$ & 100 & 17.6 \\ 
    Harp            &  3 & 31$k$ &  62$k$ & 100 & 12.0 \\
    Headphone       &  4 & 12$k$ &  25$k$ & 100 & 9.75 \\
    Windmill        &  9 & 29$k$ & 59$k$ & 100 & 13.5 \\
    \midrule
    \textbf{Total}  & 50 & – & – & 5,000 & 14.8 \\
    \bottomrule
  \end{tabular}}%
  \caption{\textbf{Statistics of our deformation benchmark.}  
  The dataset contains 50 textured objects across 10 categories. Each object is accompanied by a ground truth mesh, 100 multi‑view images, and interaction handles.}
  \label{tab-suppl-benchmark}
  \vspace{-0.6em}
\end{table}

\subsection{Ground Truth Mesh Preprocessing}
To compute ground truth of Laplacian-based deformation, we first convert each source mesh into a watertight manifold~\cite{manifold_watertight}, setting the resolution parameter to $10^5$ to preserve fine geometric detail. 
Subsequently, we apply mesh simplification~\cite{mesh_simplification} and remeshing~\cite{remesh} techniques to produce a final mesh with an average of $28k$ vertices. 
Finally, we construct a reference Laplacian operator for each processed mesh. This is computed as the standard cotangent Laplacian for triangular mesh using the NonManifold~\cite{nonmanifold}.

\section{Deformation Details} \label{suppl:deformation_detail}
We provide the details for two deformation techniques, ARAP~\cite{ARAP} and BBW~\cite{BBW} used in our experiments. For both methods, each handle is displaced by a pre-assigned translation whose magnitude is scaled to $0.2s$, where $s$ is the scale of the scene's bounding box.

\paragraph{As-Rigid-As-Possible (ARAP).} 
We utilize As-Rigid-As-Possible (ARAP)~\cite{ARAP} which manipulates the mesh by preserving the local rigidity. The method minimizes an energy defined as the sum of squared differences between edges in the deformed mesh and their corresponding rigidly rotated counterparts from the original mesh:
\begin{equation}
E(\mathbf{V}')=\sum_{i}\sum_{j\in\mathcal{N}(i)} w_{ij}\big\|\big(\mathbf{v}_i'-\mathbf{v}_j'\big)-\mathbf{R}_i\big(\mathbf{v}_i-\mathbf{v}_j\big)\big\|^2,
\label{eq:arap_energy}
\end{equation}
where $\mathbf{v}$ and $\mathbf{v}'$ are the vertices of the original and deformed meshes, respectively, $\mathbf{R}_i$ is an optimal local rotation for the neighborhood of vertex $i$, and $w_{ij}$ are the standard cotangent weights. This non-linear energy is minimized using an alternating algorithm that iterates between two steps. First, a local step finds the optimal rotations $\mathbf{R}_i$ for each vertex, and second, a global step updates all vertex positions $\mathbf{V}'$ by solving a single sparse, positive-definite linear system. 
This linear system is built from the cotangent Laplacian stiffness matrix of the underlying geometry.

Our ARAP implementation is based on the libigl~\cite{libigl}. 
To define the optimization problem and focus on local deformation evaluation, we select fixed points as all Gaussian kernels whose distance from the active handle is greater than $0.5s$.

\paragraph{Bounded Biharmonic Weights (BBW).} 
Bounded Biharmonic Weights (BBW)~\cite{BBW} is a method for generating linear blending weights that smoothly interpolate affine transformations from control handles over the entire shape.
This method finds smooth, non-negative weights $w_i$ for each control handle $i$ by solving a quadratic programming (QP) problem that minimizes a biharmonic energy. 
This energy is constructed using the discrete Laplacian stiffness matrix $L$ and mass matrix $M$, leading to the following quadratic program for each weight function $w_i$:
\begin{equation}
\min_{w_i} w_i^\top (L^\top M^{-1} L) w_i
\end{equation}
The optimization is subject to standard linear constraints, including a partition of unity ($\sum_i w_i = 1$), non-negativity ($w_i \ge 0$), and boundary conditions that fix the weights on the target vertices. The resulting weights are then used to blend handle transformations across the mesh vertices.

Specifically, we employ the Gurobi optimizer~\cite{gurobi} as a QP solver. The constraints for weights of target handles are defined by a local cage of radius 0.3$s$ around each handle, where the weights for the handle points themselves are constrained to 1 and the weights for all primitives outside the cage are constrained to 0.

\definecolor{Gray}{gray}{0.92}
\definecolor{White}{gray}{1.0}
\definecolor{LightLine}{gray}{0.7}
\arrayrulecolor{black} %

\newcolumntype{A}{>{\columncolor{Gray}\footnotesize\centering\arraybackslash}c}
\newcolumntype{B}{>{\columncolor{White}\footnotesize\centering\arraybackslash}c}

\begin{table*}[ht]
    \centering
    \small
    \setlength{\tabcolsep}{4pt} 
    \resizebox{\linewidth}{!}{
    \begin{tabular}{@{}
                    c           
                    !{\color{lightgray}\vrule}
                    A B A B A B A B A B   
                    !{\color{lightgray}\vrule}
                    A B A B A B           
                    !{\color{lightgray}\vrule}
                    A B A B               
                    !{\color{lightgray}\vrule}
                    A B                   
                    !{\color{Black}\vrule width 0.7pt}
                    A B                   
                    @{}}
        \toprule
        
        \textbf{Method}
        & \multicolumn{2}{c!{\color{LightLine}\vrule}}{\footnotesize{Bench}} & \multicolumn{2}{c!{\color{LightLine}\vrule}}{\footnotesize{Bed}} & \multicolumn{2}{c!{\color{LightLine}\vrule}}{\footnotesize{Headphone}} & \multicolumn{2}{c!{\color{LightLine}\vrule}}{\footnotesize{Basket}} & \multicolumn{2}{c!{\color{LightLine}\vrule}}{\footnotesize{Coil}}
        & \multicolumn{2}{c!{\color{LightLine}\vrule}}{\footnotesize{Birdcage}} & \multicolumn{2}{c!{\color{LightLine}\vrule}}{\footnotesize{Windmill}} & \multicolumn{2}{c!{\color{LightLine}\vrule}}{\footnotesize{Ferris Wheel}}
        & \multicolumn{2}{c!{\color{LightLine}\vrule}}{\footnotesize{Harp}} 
        & \multicolumn{2}{c!{\color{LightLine}\vrule}}{\footnotesize{Fence}}
        & \multicolumn{2}{c!{\color{Black}\vrule width 0.7pt}}{Average}
        & \multicolumn{2}{c}{\footnotesize{NS \cite{mildenhall2020nerf}}} \\ 
        \midrule
        
        \multicolumn{25}{c}{\textbf{Proxy-Based Methods}} \\
        \cmidrule(l){1-25} 
        \quad Mani‐GS \cite{gao2024mani}
        & 0.950 & 0.977 
        & 0.718 & 0.850 
        & 0.796 & 0.844 
        & 0.627 & 0.811 
        & 0.482 & 0.751
        & 0.591 & 0.807 
        & 0.744 & 0.876 
        & 0.755 & 0.909
        & 0.697 & 0.798 
        & 0.907 & 0.930 
        & 0.727 & 0.855  
        & 0.778 & 0.790
        \\
        \quad GaussianMesh \cite{MeshGaussian2024}
        & 0.952 & 0.985  
        & 0.739 & 0.872  
        & 0.795 & 0.846  
        & 0.633 & 0.841  
        & 0.482 & 0.744 
        & 0.586 & 0.806 
        & 0.729 & 0.859  
        & 0.741 & 0.889 
        & 0.665 & 0.788  
        & 0.907 & 0.931 
        & 0.723 & 0.856  
        & 0.761 & 0.784
        \\
        \quad SuGaR \cite{guedon2023sugar}
        & 0.842 & 0.590  
        & 0.779 & 0.845 
        & 0.875 & 0.631  
        & 0.874 & 0.835  
        & 0.957 & 0.779 
        & 0.902 & 0.741  
        & 0.830 & 0.831  
        & 0.806 & 0.629 
        & 0.698 & 0.589  
        & 0.795 & 0.573 
        & 0.836 & 0.704  
        & 0.834 & 0.827
        \\
        \midrule

        \multicolumn{25}{c}{\textbf{Proxy-Free Methods}} \\ \cmidrule(l){1-25}
        \quad kNN\,\scriptsize({$k$=10}) \footnotesize\cite{nonmanifold}
        & 0.585  & 0.741 
        & 0.953 & 0.957 
        & 0.931 & 0.958 
        & 0.984  & 0.952 
        & 0.838 & 0.894 
        & 0.951  & 0.963 
        & 0.927  & 0.952 
        & 0.929 & 0.952 
        & 0.860  & 0.927 
        & 0.955 & 0.944 
        & 0.891 & 0.924 
        & 0.843 & 0.872
        \\ 
        \quad kNN$^{\dag}$\,\scriptsize({$k$=30}) \footnotesize\cite{nonmanifold}
        & 0.527 & 0.700 
        & 0.901 & 0.897 
        & 0.915 & 0.967 
        & 0.970 & 0.940 
        & 0.695 & 0.746
        & 0.925 & 0.936 
        & 0.894 & 0.960 
        & 0.905 & 0.930
        & 0.627 & 0.760 
        & 0.836 & 0.861
        & 0.820 & 0.870 
        & 0.810 & 0.871
        \\
        \quad{Mahalanobis \scriptsize({$k$=10})}
        \footnotesize\cite{zhou2025laplacebeltramioperatorgaussiansplatting}
        & 0.536 & 0.784 
        & 0.937 & 0.980
        & 0.708 & 0.892 
        & 0.949 & 0.961 
        & 0.763 & 0.895 
        & 0.882 & 0.964 
        & 0.844 & 0.955 
        & 0.884 & 0.952 
        & 0.888 & 0.955 
        & 0.921 & 0.966 
        & 0.831 & 0.930 
        & 0.733 & 0.869
        \\
        \quad{Mahalanobis \scriptsize({$k$=30})}
        \footnotesize\cite{zhou2025laplacebeltramioperatorgaussiansplatting}
        & 0.522 & 0.731 
        & 0.916 & 0.981 
        & 0.704 & 0.910 
        & 0.932 & 0.951 
        & 0.746 & 0.845
        & 0.891 & 0.963 
        & 0.849 & 0.963 
        & 0.882 & 0.942
        & 0.874 & 0.938 
        & 0.894 & 0.927
        & 0.821 & 0.915 
        & 0.749 & 0.877
        \\
        \quad NeLO \cite{pang2024nelo}
         & 0.581 & 0.718 
         & 0.918 & 0.897 
         & 0.792 & 0.815 
         & 0.929 & 0.834 
         & 0.850 & 0.891
         & 0.907 & 0.922 
         & 0.878 & 0.820 
         & 0.868 & 0.882
         & 0.769 & 0.859 
         & 0.900 & 0.900
         & 0.839 & 0.854
         & 0.749 & 0.782
         \\
        \midrule
        \multicolumn{25}{@{}c}{\textbf{Ablation Result}} \\ \cmidrule(l){1-25}
        \quad{Splat Graph}
        & 0.992 & 0.994 
        & 0.976  & \textbf{0.986} 
        & 0.931  & 0.974 
        & 0.976  & \textbf{0.983} 
        & \textbf{0.980} & 0.980 
        & 0.940  & 0.991 
        & 0.909  & 0.972 
        & \textbf{0.960} & 0.970 
        & 0.842  & 0.956 
        & 0.978 & 0.993 
        & 0.948 & \textbf{0.980}
        & 0.866 & 0.883
        \\ 
        \quad\textbf{Ours}
         & \textbf{0.996} & \textbf{0.996}
         & \textbf{0.978} & 0.982
         &  \textbf{0.932} & \textbf{0.975}
         & \textbf{0.986} & 0.961
         & 0.974 & \textbf{0.990}
         & \textbf{0.986} & \textbf{0.993}
         & \textbf{0.933} & \textbf{0.973}
         & 0.946 & \textbf{0.971}
         & \textbf{0.955} & \textbf{0.966}
         & \textbf{0.991} & \textbf{0.995}
         & \textbf{0.968} & \textbf{0.980}
         & \textbf{0.878} & \textbf{0.886}
         \\
        \bottomrule
    \end{tabular}
    } 
    \caption{\textbf{Quantitative comparison of deformation quality a 3DPCK threshold of 0.05.} We report 3DPCK on our benchmark and NeRF-synthetic \cite{mildenhall2020nerf} at a threshold of 0.05 (higher is better). The score for each category is the average performance across all handles defined for objects in that category. Gray columns correspond to ARAP \cite{ARAP}, while white columns show BBW \cite{BBW} Best scores are in bold. $\dag$ indicates default parameter settings.}
    \label{suppl-tab-005}
\end{table*}
\label{suppl-tab-005}
\definecolor{Gray}{gray}{0.92}
\definecolor{White}{gray}{1.0}
\definecolor{LightLine}{gray}{0.7}
\arrayrulecolor{black} %

\newcolumntype{A}{>{\columncolor{Gray}\footnotesize\centering\arraybackslash}c}
\newcolumntype{B}{>{\columncolor{White}\footnotesize\centering\arraybackslash}c}

\begin{table*}[ht]
    \centering
    \small
    \setlength{\tabcolsep}{4pt} 
    \resizebox{\linewidth}{!}{
    \begin{tabular}{@{}
                    c           
                    !{\color{lightgray}\vrule}
                    A B A B A B A B A B   
                    !{\color{lightgray}\vrule}
                    A B A B A B           
                    !{\color{lightgray}\vrule}
                    A B A B               
                    !{\color{lightgray}\vrule}
                    A B                   
                    !{\color{Black}\vrule width 0.7pt}
                    A B                   
                    @{}}
        \toprule
        
        \textbf{Method}
        & \multicolumn{2}{c!{\color{LightLine}\vrule}}{\footnotesize{Bench}} & \multicolumn{2}{c!{\color{LightLine}\vrule}}{\footnotesize{Bed}} & \multicolumn{2}{c!{\color{LightLine}\vrule}}{\footnotesize{Headphone}} & \multicolumn{2}{c!{\color{LightLine}\vrule}}{\footnotesize{Basket}} & \multicolumn{2}{c!{\color{LightLine}\vrule}}{\footnotesize{Coil}}
        & \multicolumn{2}{c!{\color{LightLine}\vrule}}{\footnotesize{Birdcage}} & \multicolumn{2}{c!{\color{LightLine}\vrule}}{\footnotesize{Windmill}} & \multicolumn{2}{c!{\color{LightLine}\vrule}}{\footnotesize{Ferris Wheel}}
        & \multicolumn{2}{c!{\color{LightLine}\vrule}}{\footnotesize{Harp}} 
        & \multicolumn{2}{c!{\color{LightLine}\vrule}}{\footnotesize{Fence}}
        & \multicolumn{2}{c!{\color{Black}\vrule width 0.7pt}}{Average}
        & \multicolumn{2}{c}{\footnotesize{NS \cite{mildenhall2020nerf}}} \\ 
        \midrule
        
        \multicolumn{25}{c}{\textbf{Proxy-Based Methods}} \\
        \cmidrule(l){1-25} 
        \quad Mani‐GS \cite{gao2024mani}
        & 0.967 & 0.991 
        & 0.853 & 0.940 
        & 0.814 & 0.862 
        & 0.761 & 0.922 
        & 0.722 & 0.861
        & 0.816 & 0.887 
        & 0.863 & 0.922 
        & 0.863 & 0.941
        & 0.843 & 0.839 
        & 0.960 & 0.960 
        & 0.846 & 0.913 
        & 0.839 & 0.848
        \\
        \quad GaussianMesh \cite{MeshGaussian2024}
        & 0.968 & 0.993  
        & 0.857 & 0.941  
        & 0.806 & 0.860  
        & 0.765 & 0.925  
        & 0.711 & 0.859 
        & 0.816 & 0.883 
        & 0.857 & 0.912  
        & 0.854 & 0.936 
        & 0.797 & 0.887  
        & 0.960 & 0.959 
        & 0.839 & 0.916  
        & 0.831 & 0.846
        \\
        \quad SuGaR \cite{guedon2023sugar}
        & 0.883 & 0.695  
        & 0.891 & 0.925 
        & 0.911 & 0.769  
        & 0.959 & 0.910  
        & 0.984 & 0.869 
        & 0.974 & 0.830  
        & 0.934 & 0.901  
        & 0.894 & 0.744 
        & 0.872 & 0.688  
        & 0.894 & 0.693 
        & 0.920 & 0.802  
        & 0.898 & 0.881
        \\
        \midrule

        \multicolumn{25}{c}{\textbf{Proxy-Free Methods}} \\ \cmidrule(l){1-25}
        \quad kNN\,\scriptsize({$k$=10}) \footnotesize\cite{nonmanifold}
        & 0.710  & 0.819 
        & 0.992 & 0.990 
        & \textbf{0.955} & \textbf{0.984}
        & \textbf{0.995}  & 0.980 
        & 0.937 & 0.950 
        & 0.980  & 0.980 
        & 0.988  & 0.983 
        & 0.969 & 0.982 
        & 0.947  & 0.965 
        & 0.992 & 0.967 
        & 0.947 & 0.960 
        & 0.941 & 0.933
        \\ 
        \quad kNN$^{\dag}$\,\scriptsize({$k$=30}) \footnotesize\cite{nonmanifold}
        & 0.663 & 0.792 
        & 0.970 & 0.966 
        & \textbf{0.955} & 0.983 
        & 0.988 & 0.975 
        & 0.835 & 0.841
        & 0.986 & 0.967 
        & 0.976 & 0.985 
        & 0.961 & 0.968
        & 0.788 & 0.845 
        & 0.954 & 0.918
        & 0.908 & 0.924 
        & 0.936 & 0.931
        \\
        \quad{Mahalanobis \scriptsize({$k$=10})}
        \footnotesize\cite{zhou2025laplacebeltramioperatorgaussiansplatting}
        & 0.721 & 0.866 
        & 0.992 & \textbf{0.997} 
        & 0.804 & 0.946 
        & 0.987 & 0.988 
        & 0.922 & 0.948 
        & 0.969 & 0.992 
        & 0.952 & 0.984 
        & 0.967 & 0.985 
        & 0.965 & 0.976 
        & 0.992 & 0.995 
        & 0.927 & 0.968 
        & 0.908 & 0.935
        \\
        \quad{Mahalanobis \scriptsize({$k$=30})}
        \footnotesize\cite{zhou2025laplacebeltramioperatorgaussiansplatting}
        & 0.676 & 0.828 
        & 0.979 & 0.995 
        & 0.794 & 0.949 
        & 0.983 & 0.983 
        & 0.906 & 0.908
        & 0.977 & 0.986 
        & 0.955 & 0.988 
        & 0.963 & 0.984
        & 0.963 & 0.969 
        & 0.989 & 0.981
        & 0.919 & 0.957 
        & 0.914 & 0.936
        \\
        \quad NeLO \cite{pang2024nelo}
         & 0.737 & 0.823 
         & 0.988 & 0.962 
         & 0.904 & 0.927 
         & 0.980 & 0.918 
         & 0.948 & 0.956
         & 0.972 & 0.966 
         & 0.962 & 0.909 
         & 0.959 & 0.945
         & 0.952 & 0.943 
         & 0.977 & 0.964
         & 0.938 & 0.931 
         & 0.910 & 0.902
         \\
        \midrule
        \multicolumn{25}{@{}c}{\textbf{Ablation Result}} \\ \cmidrule(l){1-25}
        \quad{Splat Graph}
        & 0.997 & 0.997 
        & 0.996  & 0.994 
        & \textbf{0.955}  & 0.982 
        & 0.993  & \textbf{0.990} 
        & \textbf{0.998} & 0.988 
        & 0.991  & 0.996 
        & 0.984  & 0.981 
        & \textbf{0.984} & \textbf{0.988} 
        & 0.939  & 0.971 
        & 0.999 & 0.997 
        & 0.983 & 0.988 
        & 0.947 & 0.943
        \\ 
        \quad\textbf{Ours}
         & \textbf{0.998} & \textbf{0.999}
         & \textbf{0.997} & \textbf{0.997}
         &  \textbf{0.955} & \textbf{0.984}
         & \textbf{0.995} & 0.984
         & 0.988 & \textbf{0.998}
         & \textbf{0.998} & \textbf{0.998}
         & \textbf{0.993} & \textbf{0.993}
         & 0.968 & \textbf{0.988}
         & \textbf{0.992} & \textbf{0.987}
         & \textbf{1.000} & \textbf{1.000}
         & \textbf{0.988} & \textbf{0.992} 
         & \textbf{0.949} & \textbf{0.945}
         \\
        \bottomrule
    \end{tabular}
    } 
    \caption{\textbf{Quantitative comparison of deformation quality at a 3DPCK threshold of 0.1.} We report 3DPCK on our benchmark and NeRF-synthetic \cite{mildenhall2020nerf} at a threshold of 0.1 (higher is better). The score for each category is the average performance across all handles defined for objects in that category. Gray columns correspond to ARAP \cite{ARAP}, while white columns show BBW \cite{BBW} Best scores are in bold. $\dag$ indicates default parameter settings.}
    \label{suppl-tab-01}
\end{table*}
\label{suppl-tab-01}

\section{Benchmark Details}\label{suppl-benchmark-section}
Here, we provide further details on our benchmark. This benchmark includes 50 textured objects spanning 10 categories, chosen specifically to cover topologically challenging scenarios such as object including thin details or adjacent but distinct surfaces. 
The models were sourced from ShapeNet~\cite{ShapeNet}, Objaverse~\cite{objaverse} and SketchFab~\cite{Pinson2011Sketchfab}. 
The objects sourced from SketchFab were used under licenses that permit academic use.
All objects in our benchmark are normalized to range of [-1.3, 1.3], resulting in a consistent scene scale $s$ of 2.6.
For each object, we provide a high quality ground truth mesh, and a set of 100 multi-view images rendered at $1920\times1080$ resolution and interaction handle annotations. 
The handles, averaging 14.8 per object, were placed by human annotators in regions known to be susceptible to deformation artifacts. 
These locations include joints, thin parts, and areas where distinct surfaces are close, creating challenging test cases for neighborhood estimation.
Tab.~\ref{tab-suppl-benchmark} provides a detailed statistical breakdown of the benchmark. Fig.~\ref{fig-suppl-benchmark-vis} visualizes example objects included in the benchmark.

\section{Additional Experiments}
\label{sec:supl_more_results}
\subsection{Analysis on the 3DPCK Threshold}\label{suppl-varying-tau}
In our main experiments, we report 3DPCK scores with a threshold of $0.075$. 
We use a moderate threshold of $0.075$ to ensure a fair evaluation that is neither overly sensitive to noise nor overly permissive of large errors.

To validate our comparison across different evaluation thresholds, we provide a quantitative comparison with varying thresholds in Tab.~\ref{suppl-tab-005},~\ref{suppl-tab-01}. 
We tested with threshold $0.05$ (stricter case) and $0.1$ (looser case), and found that the overall performance trend, with our method consistently ranking best, remained the same. 

Note that, because keypoints are sampled from the ground truth mesh, our 3DPCK evaluation is inherently robust against incorrectly located but transparent kernels. 
Furthermore, since the reconstructed Gaussian kernels are sufficiently dense, the metric is also unaffected by initial registration errors, ensuring it isolates the quality of the deformation itself.
The rare cases of pairing failure were exclusively due to reconstruction errors caused by defects in the source mesh; since these errors directly impact the final rendering quality, they are correctly penalized by our metric.

\subsection{Effect of Varying Neighborhood Size $k$} \label{suppl-effect-hyperparameter}
\begin{figure*}[!t]
    \centering
    \setlength{\tabcolsep}{3pt}
    \resizebox{\linewidth}{!}{%
    \begin{tabular}{@{}
        >{\centering\arraybackslash}m{0.02\linewidth}
        >{\centering\arraybackslash}m{0.03\linewidth}
        >{\centering\arraybackslash}m{0.18\linewidth}|
        >{\centering\arraybackslash}m{0.18\linewidth}
        >{\centering\arraybackslash}m{0.18\linewidth}
        >{\centering\arraybackslash}m{0.18\linewidth}|
        >{\centering\arraybackslash}m{0.18\linewidth}
        >{\centering\arraybackslash}m{0.18\linewidth}
        >{\centering\arraybackslash}m{0.18\linewidth}
        >{\centering\arraybackslash}m{0.18\linewidth}
    @{}}
    
    & & 
    \multicolumn{1}{>{\centering\arraybackslash}m{0.18\linewidth}}{\multirow{2}{*}{\large\textbf{Source}}} & 
    \multicolumn{3}{c}{\textbf{SpLap}} & 
    \multicolumn{3}{c}{\textbf{Euclidean Distance \cite{nonmanifold}}}  \\
    \cmidrule(lr){4-6} \cmidrule(lr){7-9}
    
    & & & 
    \textbf{$k\!=10$} & 
    \textbf{$k\!=30^\dagger$} & 
    \textbf{$k\!=50$} & 
    \textbf{$k\!=10^\dagger$} & 
    \textbf{$k\!=30^\ddagger$} & 
    \textbf{$k\!=50$} \\
    \toprule

    \multirow{3}{*}{\large\adjustbox{center, right=-6.4cm, angle=90, valign=m}{\textsc{As-Rigid-As-Possible \cite{ARAP}}}}
    & \adjustbox{center, right=0.25cm, angle=90, valign=m}{\textbf{Birdcage}}
    & \includegraphics[width=\linewidth, valign=m]{images/main_qual/GT/Birdcage/arrow.png}
    & \includegraphics[width=\linewidth, valign=m]{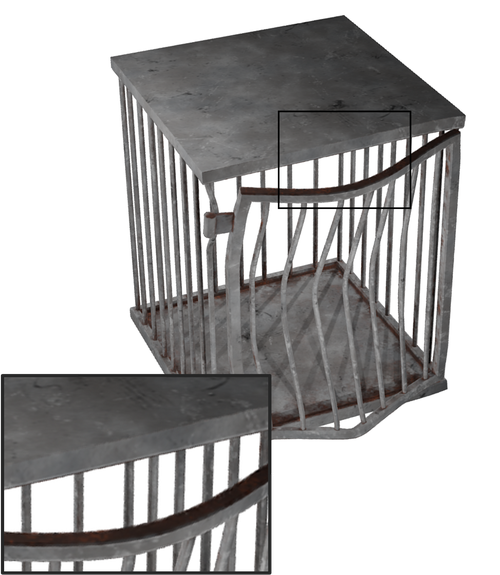}
    & \includegraphics[width=\linewidth, valign=m]{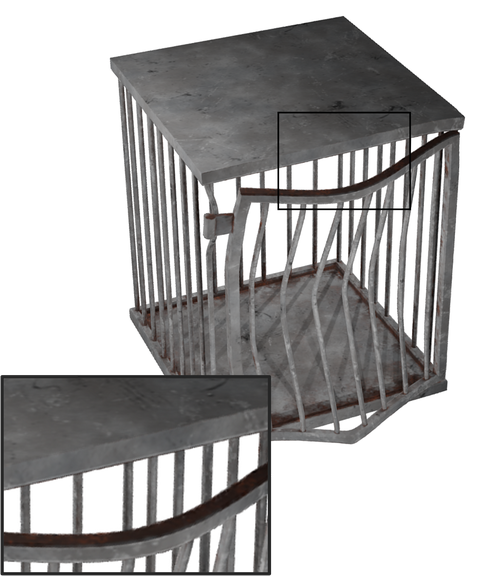}
    & \includegraphics[width=\linewidth, valign=m]{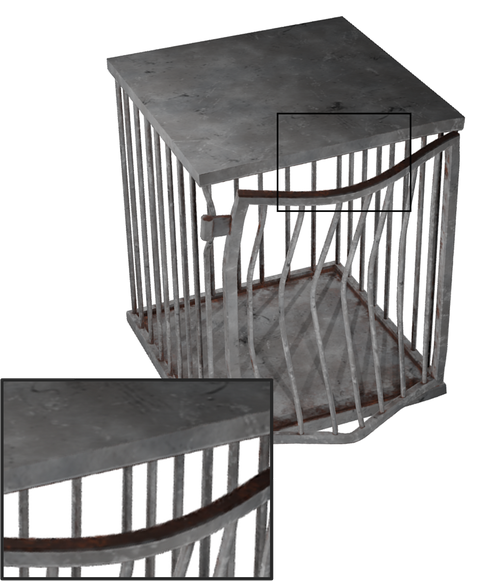}
    & \includegraphics[width=\linewidth, valign=m]{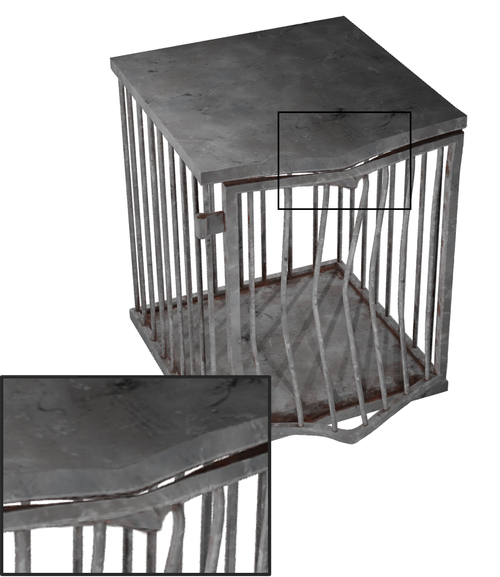}
    & \includegraphics[width=\linewidth, valign=m]{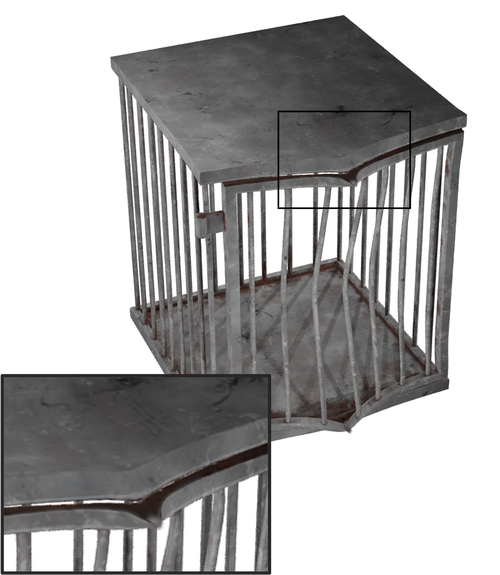}
    & \includegraphics[width=\linewidth, valign=m]{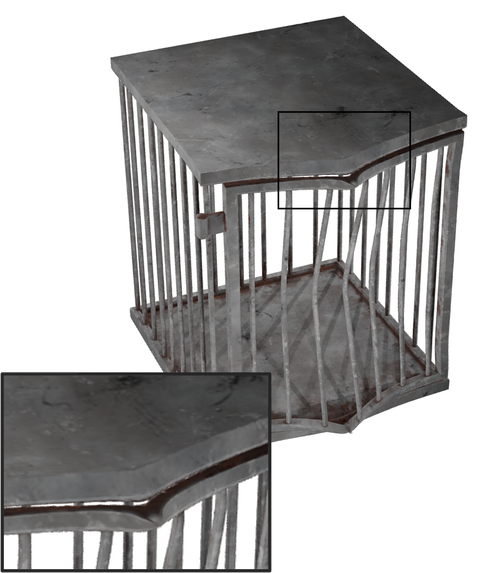} \\ 
    
    & \adjustbox{center, right=0.3cm, angle=90, valign=m}{\textbf{Harp}}
    & \includegraphics[width=\linewidth, valign=m]{images/main_qual/harp-arap/gt/arrow.png}
    & \includegraphics[width=\linewidth, valign=m]{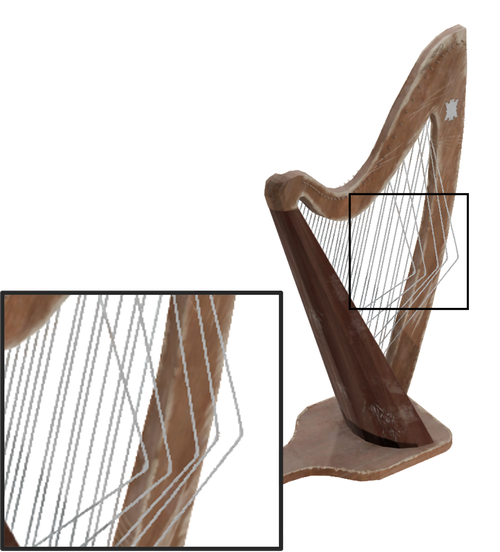}
    & \includegraphics[width=\linewidth, valign=m]{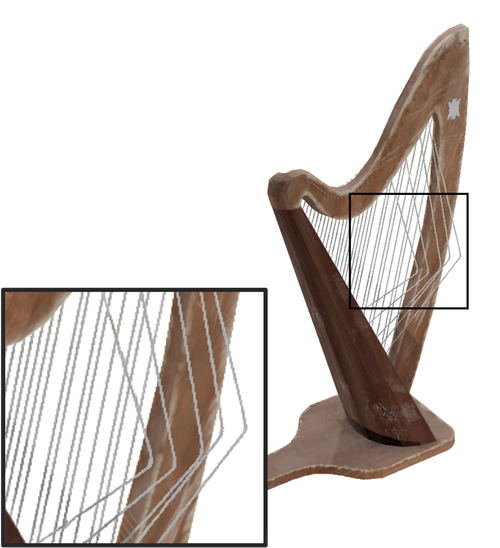}
    & \includegraphics[width=\linewidth, valign=m]{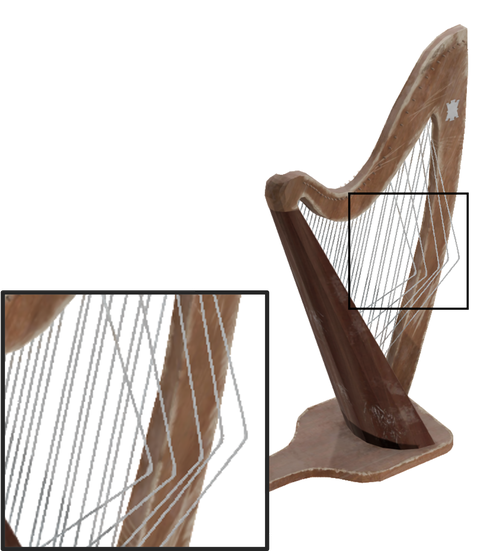}
    & \includegraphics[width=\linewidth, valign=m]{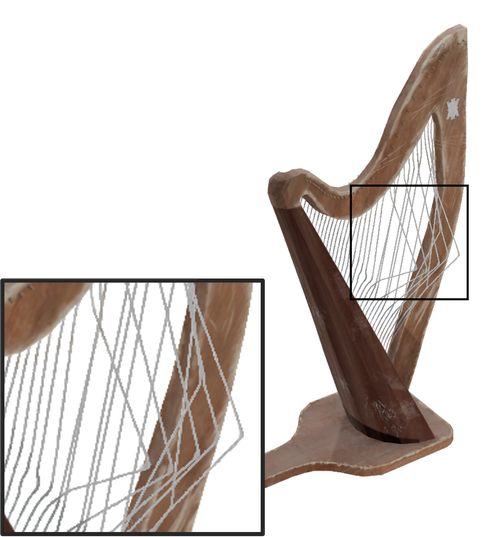}
    & \includegraphics[width=\linewidth, valign=m]{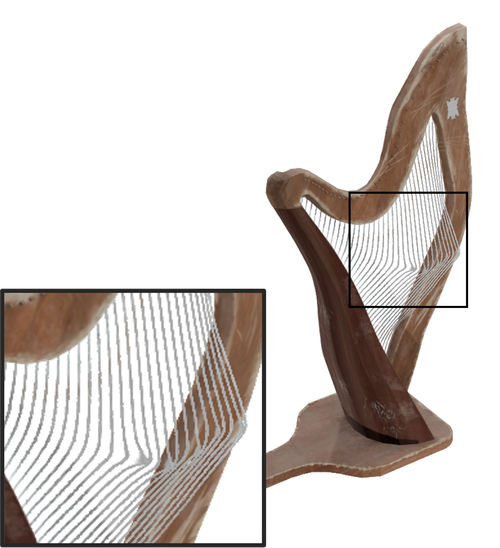}
    & \includegraphics[width=\linewidth, valign=m]{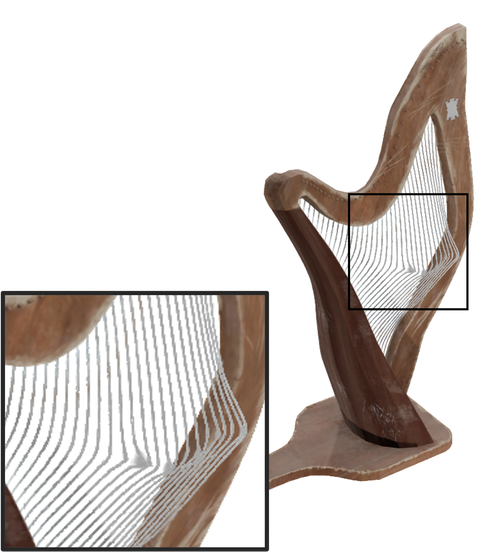} \\ [3pt]
    
    & \adjustbox{center, right=0.3cm, angle=90, valign=m}{\textbf{Windmill}}
    & \includegraphics[width=\linewidth, valign=m]{images/main_qual/GT/Windmill/arrow.png}
    & \includegraphics[width=\linewidth, valign=m]{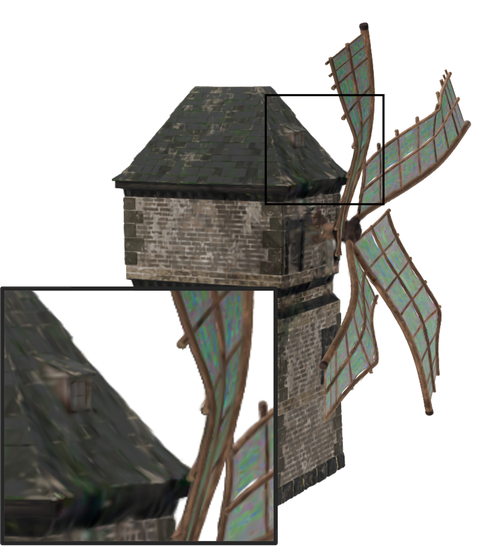}
    & \includegraphics[width=\linewidth, valign=m]{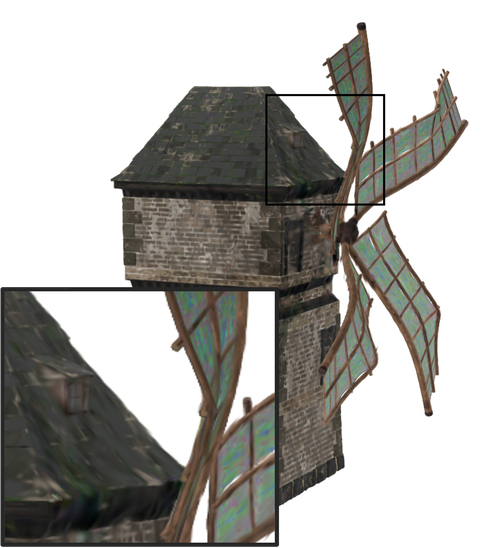}
    & \includegraphics[width=\linewidth, valign=m]{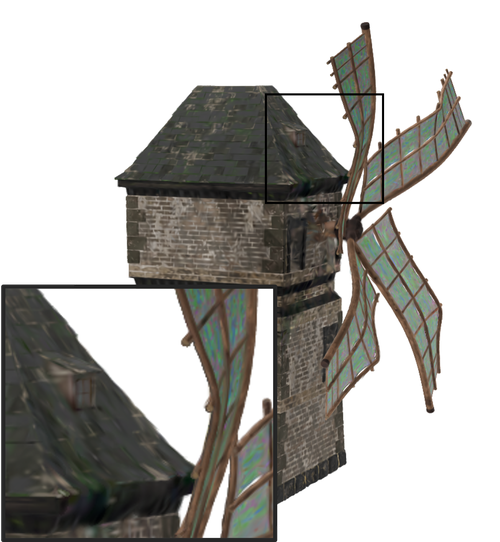}
    & \includegraphics[width=\linewidth, valign=m]{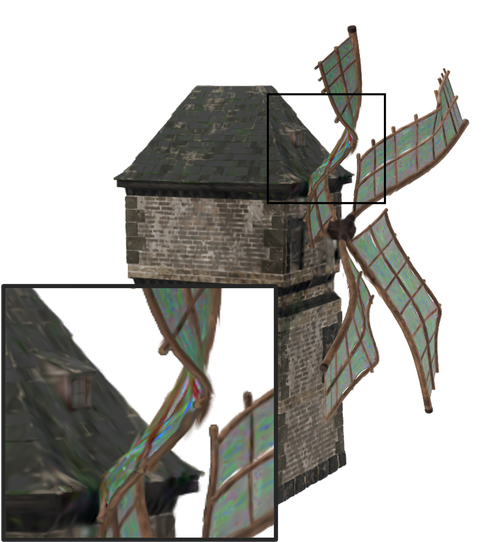}
    & \includegraphics[width=\linewidth, valign=m]{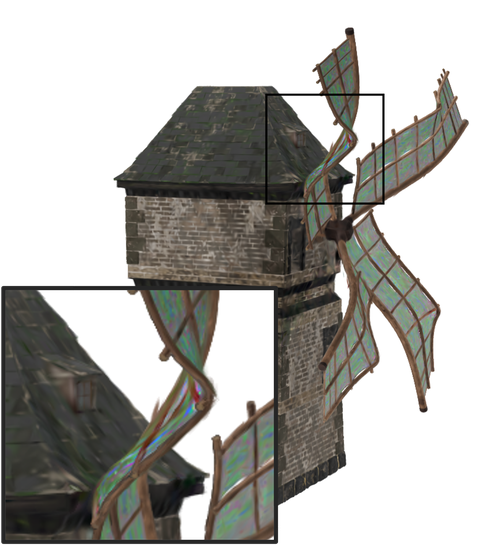}
    & \includegraphics[width=\linewidth, valign=m]{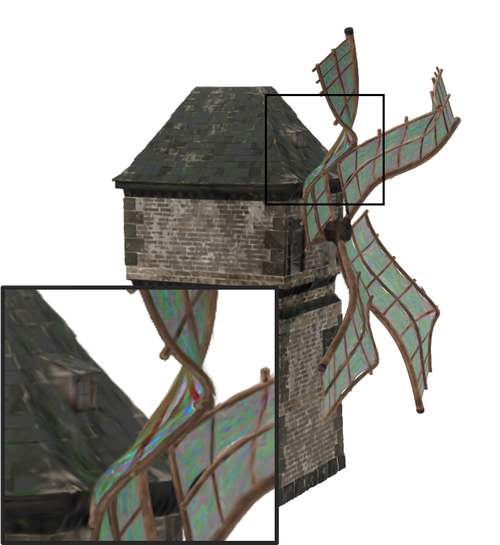} \\

    & \adjustbox{center, right=0.3cm, angle=90, valign=m}
    {\textbf{Bench 1}}
    & \includegraphics[width=\linewidth, valign=m]{images/main_qual/bench-bbw/gt/arrow.png}
    & \includegraphics[width=\linewidth, valign=m]{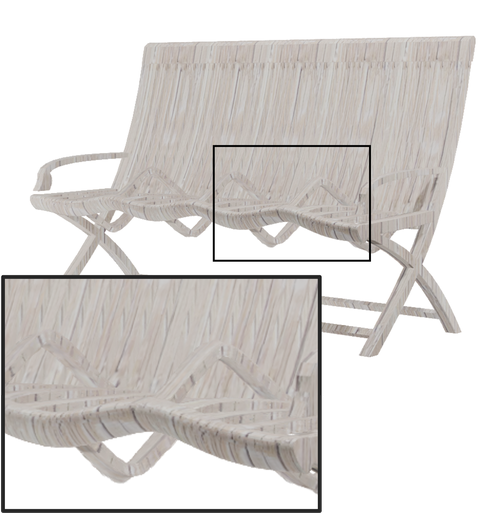}
    & \includegraphics[width=\linewidth, valign=m]{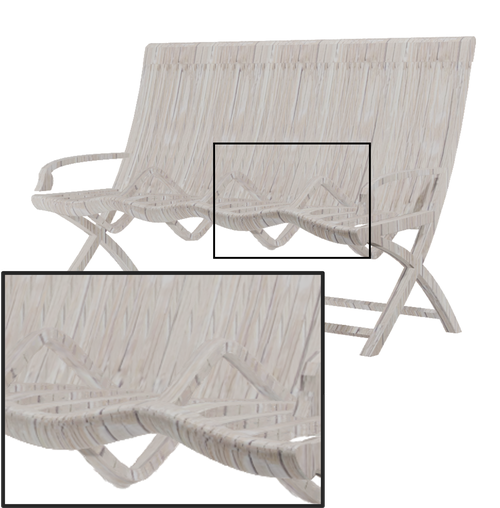}
    & \includegraphics[width=\linewidth, valign=m]{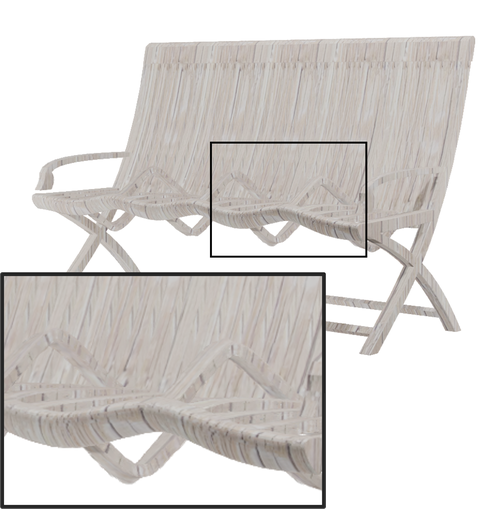}
    & \includegraphics[width=\linewidth, valign=m]{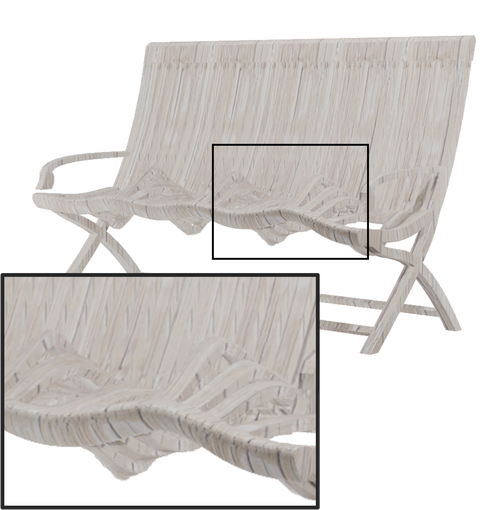}
    & \includegraphics[width=\linewidth, valign=m]{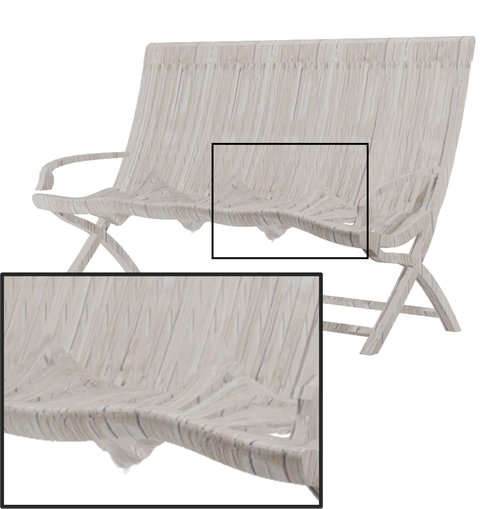}
    & \includegraphics[width=\linewidth, valign=m]{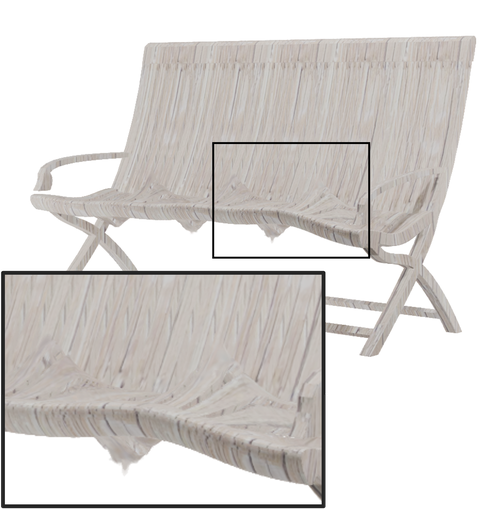} 
    \\ [5pt]

    & \adjustbox{center, right=0.25cm, angle=90, valign=m}
    {\textbf{Bed}}
    & \includegraphics[width=\linewidth, valign=m]{images/main_qual/bed-arap/gt/arrow.png}
    & \includegraphics[width=\linewidth, valign=m]{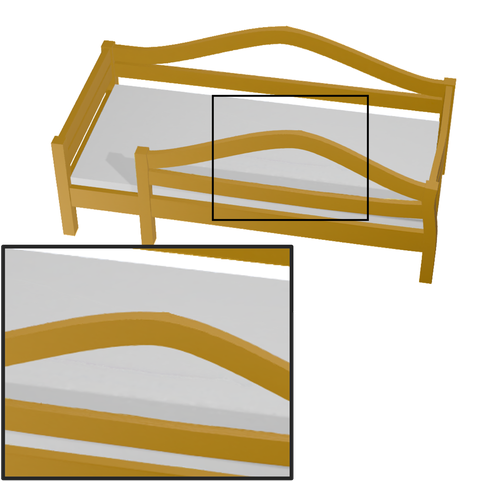}
    & \includegraphics[width=\linewidth, valign=m]{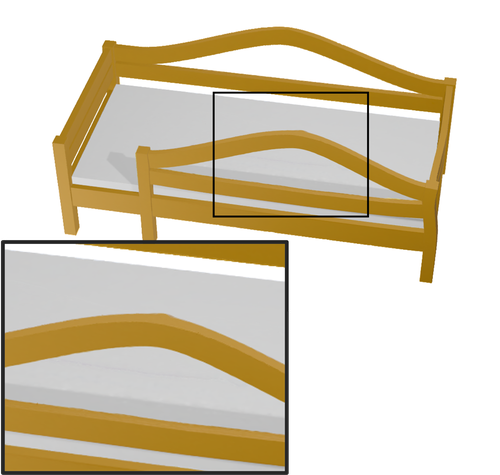}
    & \includegraphics[width=\linewidth, valign=m]{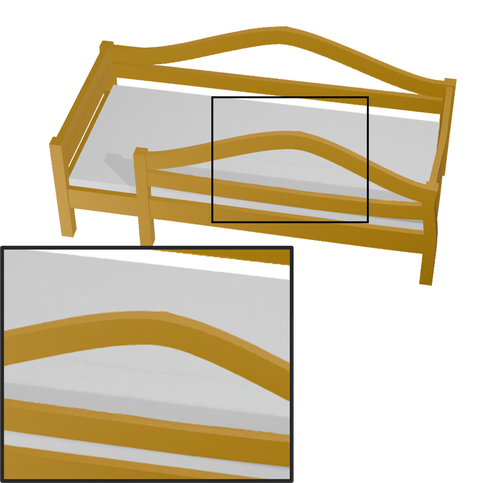}
    & \includegraphics[width=\linewidth, valign=m]{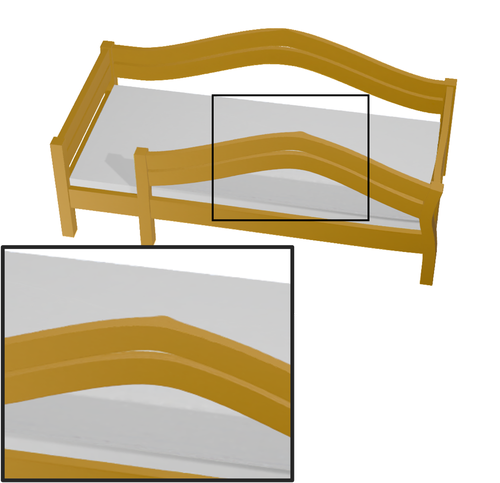}
    & \includegraphics[width=\linewidth, valign=m]{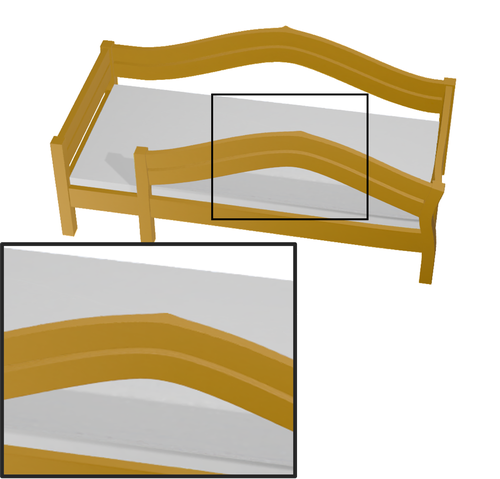}
    & \includegraphics[width=\linewidth, valign=m]{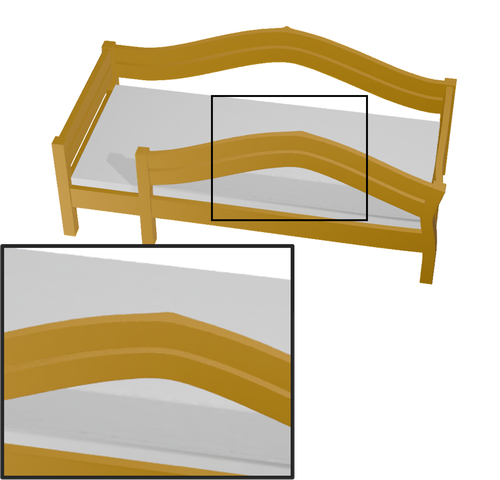}  
    \\ [7pt]
    \midrule
    \multirow{2}{=}{\large\adjustbox{center, right=-3.2cm, angle=90, valign=m}{\textsc{Bounded Biharmonic Weights \cite{BBW}}}}

    & \adjustbox{center, right=0.3cm, angle=90, valign=m}{\textbf{Coil}}
    & \includegraphics[width=\linewidth, valign=m]{images/main_qual/GT/Coil/arrow.png}
    & \includegraphics[width=\linewidth, valign=m]{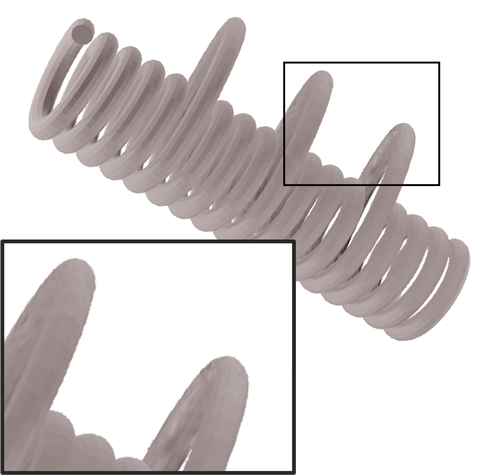}
    & \includegraphics[width=\linewidth, valign=m]{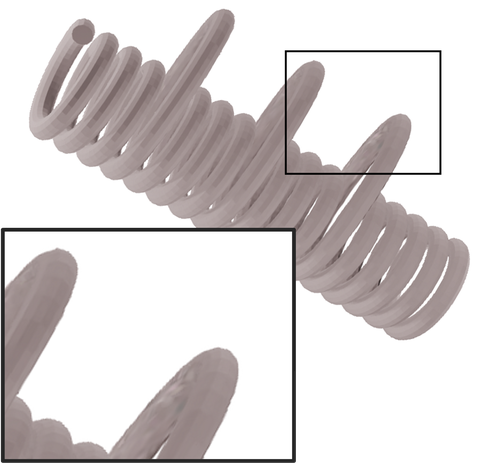}
    & \includegraphics[width=\linewidth, valign=m]{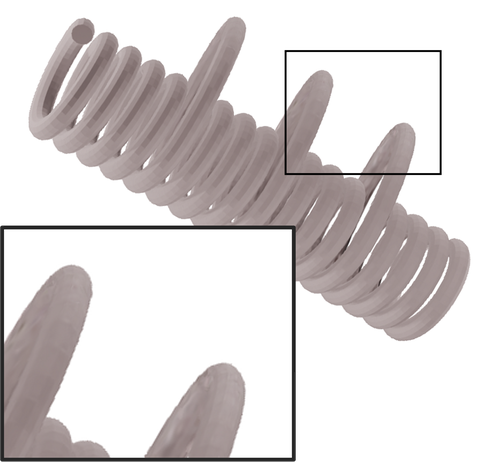}
    & \includegraphics[width=\linewidth, valign=m]{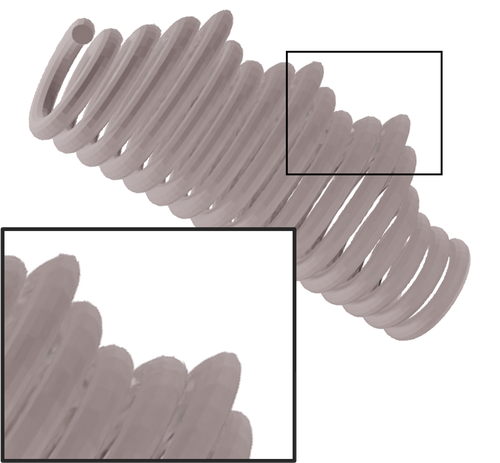}
    & \includegraphics[width=\linewidth, valign=m]{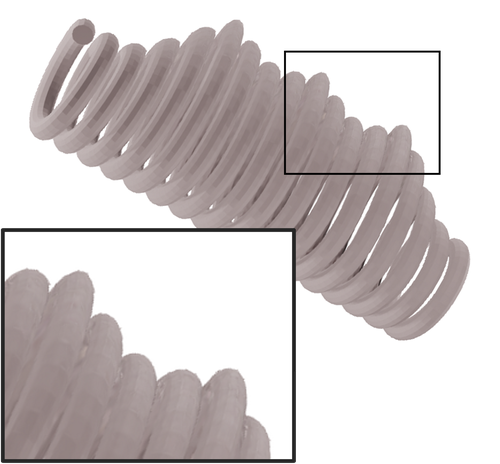}
    & \includegraphics[width=\linewidth, valign=m]{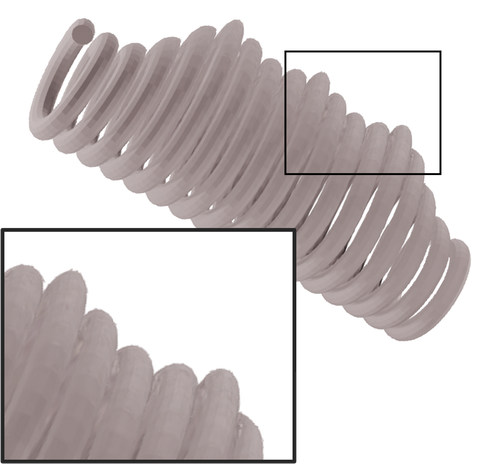} \\ [3pt]
    
    & \adjustbox{center, right=0.25cm, angle=90, valign=m}{\textbf{Bench 2}}
    & \includegraphics[width=\linewidth, valign=m]{images/main_qual/GT/Bench/arrow.png}
    & \includegraphics[width=\linewidth, valign=m]{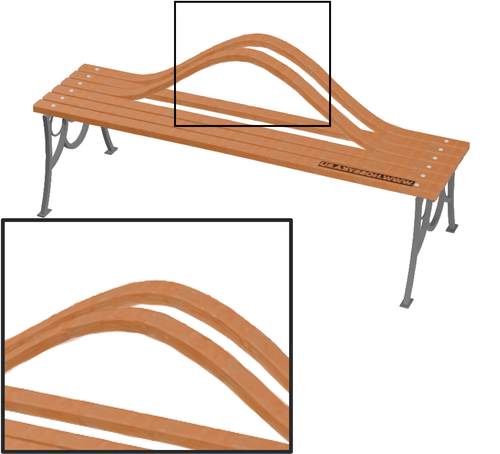}
    & \includegraphics[width=\linewidth, valign=m]{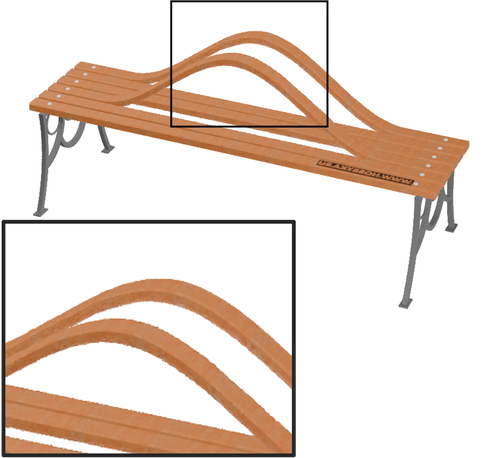}
    & \includegraphics[width=\linewidth, valign=m]{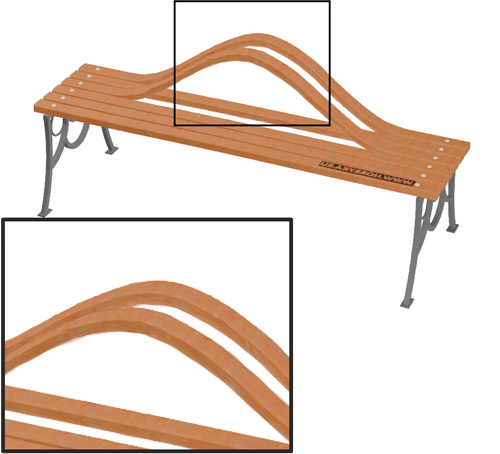}
    & \includegraphics[width=\linewidth, valign=m]{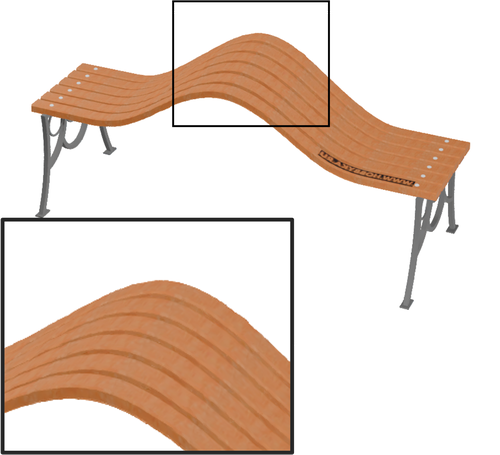}
    & \includegraphics[width=\linewidth, valign=m]{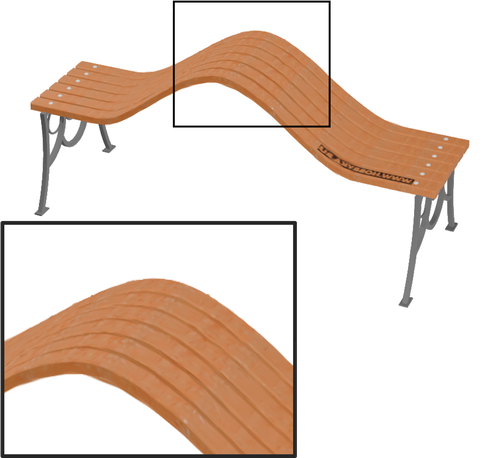}
    & \includegraphics[width=\linewidth, valign=m]{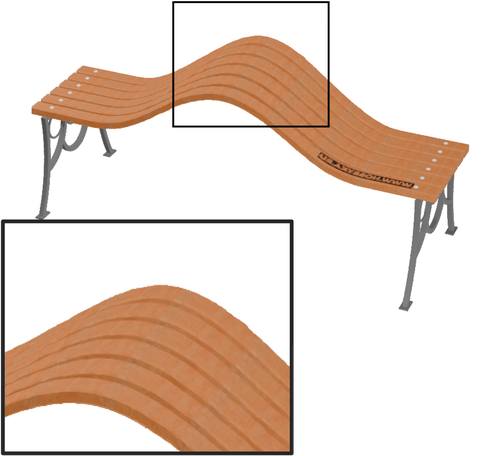} \\ [5pt]
    
    & \adjustbox{center, right=0.25cm, angle=90, valign=m}{\textbf{Fence}}
    & \includegraphics[width=\linewidth, valign=m]{images/main_qual/fence-bbw/gt/arrow.png}
    & \includegraphics[width=\linewidth, valign=m]{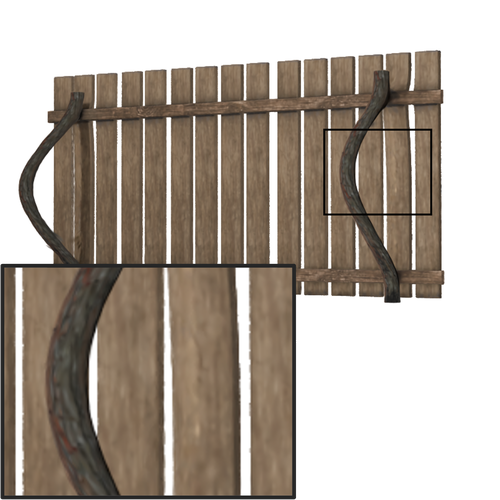}
    & \includegraphics[width=\linewidth, valign=m]{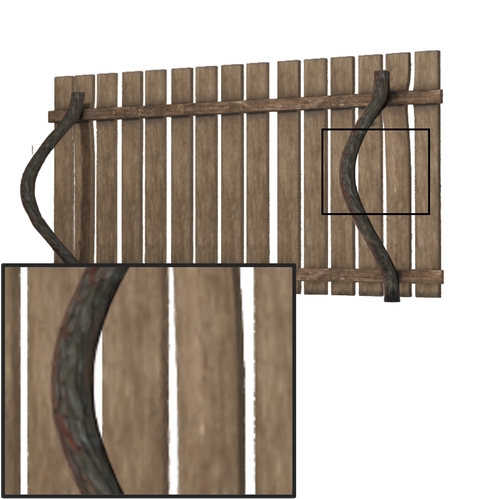}
    & \includegraphics[width=\linewidth, valign=m]{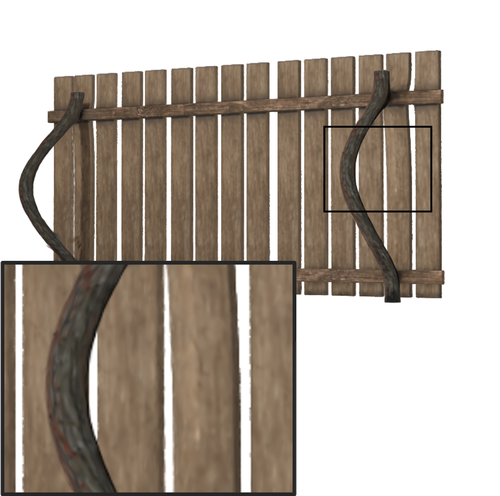}
    & \includegraphics[width=\linewidth, valign=m]{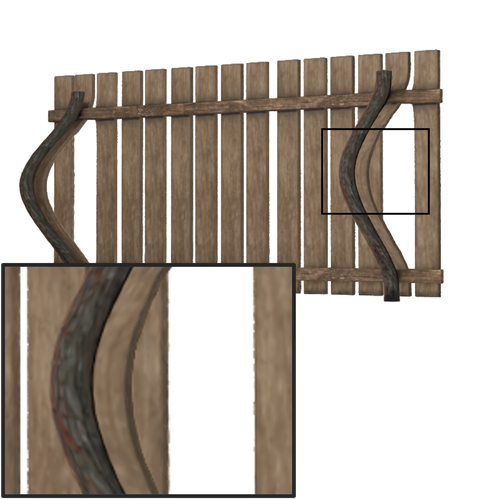}
    & \includegraphics[width=\linewidth, valign=m]{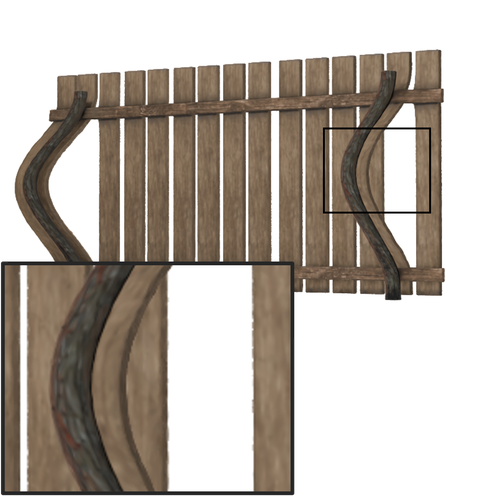}
    & \includegraphics[width=\linewidth, valign=m]{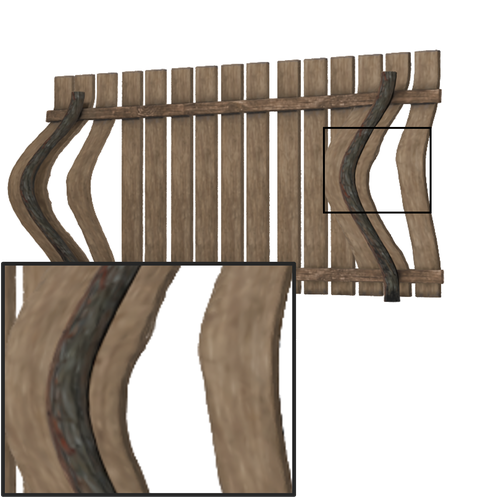} \\
    
    \\
    \bottomrule
    \end{tabular}
    }
    \caption{
    \textbf{Qualitative analysis of hyperparameter sensitivity.}
    The red dots denote the interaction handles, with arrows indicating the edit direction.
    The top five rows show results using the ARAP \cite{ARAP}, while the bottom three rows use BBW \cite{BBW}. 
    $\dagger$ denotes the optimal parameter setting used in our main experiments. $\ddagger$ denotes the default parameter setting. Our method demonstrates superior robustness to the choice of the number of neighbors $k$, as it leverages intrinsic surface distance rather than simple spatial proximity.
    }
    \label{fig:suppl_abl_qual}
\end{figure*}

We demonstrate our method's robustness to hyperparameter selection with a qualitative analysis in Fig.~\ref{fig:suppl_abl_qual}. 
Conventional geometry estimation for point clouds typically requires meticulous tuning of the hyperparameter $k$, the number of neighbors. 
This presents a critical challenge, as a $k$ that is too small can induce graph fragmentation, while a $k$ that is too large often causes the neighborhood to expand off the surface, leading to incorrect connections. 
Both scenarios result in severe visual artifacts during deformation. 
This trade-off becomes particularly acute for objects with intricate geometry, where a single value for $k$ that avoids both failure modes may not exist.

In contrast, our framework is fundamentally more robust because it is based on the intrinsic distance along the surface manifold, not simple spatial proximity. As shown in Fig.~\ref{fig:suppl_abl_qual}, our method maintains high quality results across a wide range of values, from $k\!=\!10$ to an aggressively large $k\!=\!50$, whereas the conventional Euclidean-based method clearly fails as $k$ increases. 
This demonstrates not only the robustness of our method to this critical hyperparameter, but also that superior performance of our method is a result of its fundamental design, not merely an artifact of careful parameter tuning.

\begin{figure*}[t]
  \centering
  \includegraphics[width=\textwidth]{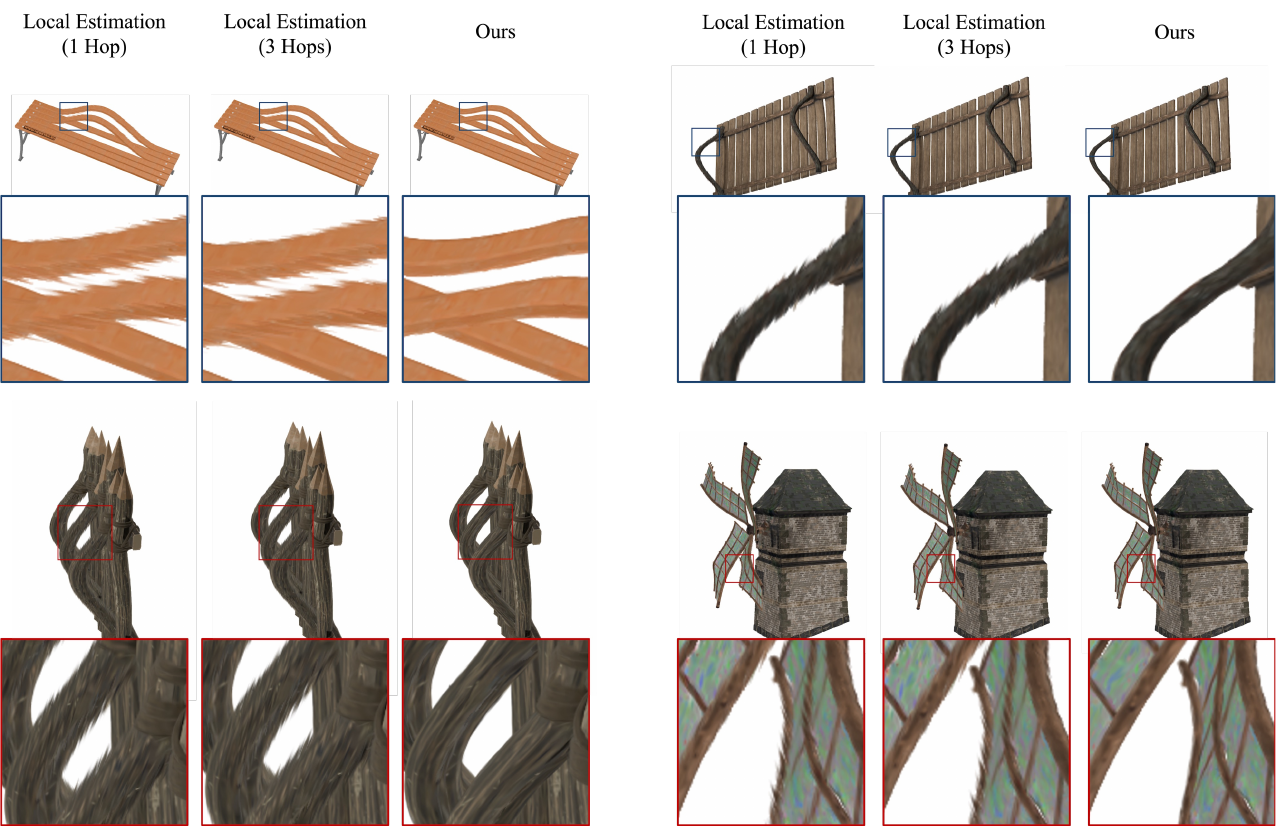}
  \caption{
  \textbf{Comparison of different kernel adaptation methods.}
$N$ Hop(s) denotes the local transformation estimation based on neighbors within $N$
edges. While simple local transformations fail even with wider neighborhoods (3 hops), our method effectively adapts each kernel to maintain high visual quality.}
  \label{fig:suppl_adap_more_result}
\end{figure*}

\subsection{Additional Results of Kernel Adaptation Ablation}
Fig.~\ref{fig:suppl_adap_more_result} presents additional qualitative results comparing our kernel adaptation against a simple local affine transformation estimation described in Surface-Preserving Adaptation Section in main paper. 
These examples further illustrate the failure modes of the naive approach. 
Due to the fundamental scale mismatch between the Gaussian kernel and corresponding local neighborhood, the estimated transformation fails to accurately model the underlying surface, particularly under significant bending or large-scale deformations.

In contrast, our method is significantly more robust. By directly adapting each kernel to maintain its coverage on the deformed surface manifold, our approach preserves geometric consistency and visual fidelity even in these challenging cases.


\end{document}